\documentclass[journal]{IEEEtran}
\usepackage{amsmath,amsfonts}
\usepackage{algorithm}
\usepackage{algpseudocode}
\usepackage{multirow}
\usepackage{pifont}
\usepackage{tabularx}
\usepackage{array}
\usepackage[caption=false,font=normalsize,labelfont=sf,textfont=sf]{subfig}
\usepackage{textcomp}
\usepackage{stfloats}
\usepackage{color}
\usepackage{url}
\usepackage{verbatim}
\usepackage{graphicx}
\usepackage{cite}
\usepackage{hyperref}

\usepackage[table,xcdraw]{xcolor}

\begin{document}

\title{Progressive Boundary Guided Anomaly Synthesis for Industrial Anomaly Detection}

\author{
Qiyu Chen, Huiyuan Luo, Han Gao, Chengkan Lv,~\IEEEmembership{Member,~IEEE}, Zhengtao Zhang,~\IEEEmembership{Member,~IEEE}

\thanks{Manuscript created 16 June 2024.
This work was supported in part by the National Natural Science Foundation of China under Grant No. 62303458 and 62303461.
\textit{(Corresponding author: Chengkan Lv.)}

Qiyu Chen, Huiyuan Luo, Han Gao, Chengkan Lv, and Zhengtao Zhang are
with the Institute of Automation, Chinese Academy of Sciences, Beijing 100190, China,
and also with the School of Artificial Intelligence, University of Chinese Academy of Sciences, Beijing 100049, China
(e-mail: chenqiyu2021@ia.ac.cn, huiyuan.luo@ia.ac.cn, gaohan2022@ia.ac.cn,
chengkan.lv@ia.ac.cn, zhengtao.zhang@ia.ac.cn).}
}

\maketitle

\begin{abstract}
Unsupervised anomaly detection methods can identify surface defects in industrial images by leveraging only normal samples for training.
Due to the risk of overfitting when learning from a single class,
anomaly synthesis strategies are introduced to enhance detection capability by generating artificial anomalies.
However, existing strategies heavily rely on anomalous textures from auxiliary datasets.
Moreover, their limitations in the coverage and directionality of anomaly synthesis
may result in a failure to capture useful information and lead to significant redundancy.
To address these issues, we propose a novel Progressive Boundary-guided Anomaly Synthesis (PBAS) strategy,
which can directionally synthesize crucial feature-level anomalies without auxiliary textures.
It consists of three core components: Approximate Boundary Learning (ABL),
Anomaly Feature Synthesis (AFS), and Refined Boundary Optimization (RBO).
To make the distribution of normal samples more compact,
ABL first learns an approximate decision boundary by center constraint,
which improves the center initialization through feature alignment.
AFS then directionally synthesizes anomalies with more flexible scales guided by the hypersphere distribution of normal features.
Since the boundary is so loose that it may contain real anomalies, 
RBO refines the decision boundary through the binary classification of artificial anomalies and normal features.
Experimental results show that our method achieves state-of-the-art performance and the fastest detection speed
on three widely used industrial datasets, including MVTec AD, VisA, and MPDD.
The code will be available at: \url{https://github.com/cqylunlun/PBAS}.
\end{abstract}

\section{Introduction}
\label{sec:intro}

\IEEEPARstart{A}{nomaly} detection aims to identify unseen data points that deviate from the
normal data distribution.
Recently, it has been widely applied in various domains, including industrial inspection \cite{roth2022towards,luo2024template,yao2023scalable},
medical diagnosis \cite{zhou2021proxy,guo2023encoder,lu2024anomaly},
and video surveillance \cite{zhong2022bidirectional,zeng2021hierarchical,zhang2022influence}.
In the field of industrial inspection, anomalies typically refer to various types of surface defects on products, such as scratches, cracks, and stains.
However, it is challenging to collect all defect patterns in real-world applications for supervised learning.
Additionally, the cost of precise pixel-level annotations for guiding anomaly localization is prohibitively high.
Therefore, unsupervised anomaly detection (UAD) is crucial for identifying defective products in manufacturing processes.

The UAD methods leverage defect-free images to train models, which can be broadly classified into three categories.
Reconstruction-based methods \cite{zavrtanik2021reconstruction,pirnay2022inpainting,luo2024ami} aim to reconstruct input images
from the latent space and detect anomalies by analyzing the reconstruction error.
However, these methods heavily rely on the quality of reconstructed images,
which consequently faces challenges in difference analysis.
Embedding-based methods \cite{lee2022cfa,deng2022anomaly,liu2023unsupervised} aim to learn more distinctive embeddings for input images
and detect anomalies by measuring the distance between input data and learned representations.
Despite achieving superior performance, these models have only been exposed to normal samples.

\begin{figure}[t]
    \centering
    \includegraphics[width=0.99\linewidth]{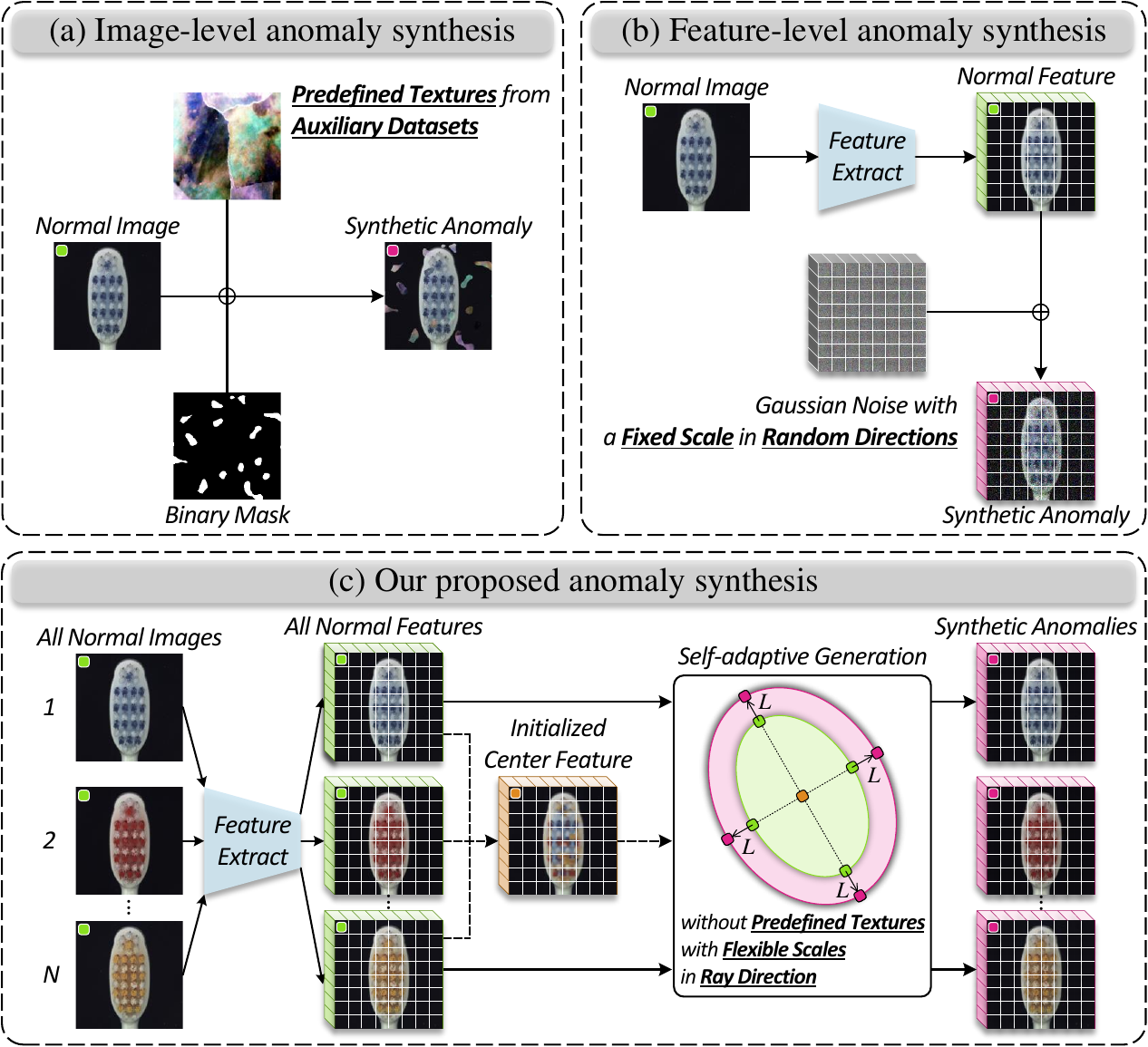}
    \caption{Overview of the existing and proposed anomaly synthesis strategies.
    (a) Image-level anomaly synthesis strategies heavily rely on the predefined textures from auxiliary datasets.
    (b) Feature-level anomaly synthesis strategies utilize Gaussian noise with a fixed scale in random directions.
    (c) Our proposed method directionally synthesizes feature-level anomalies with flexible scales and without predefined properties.
    }
    \label{fig:category}
\end{figure}

To mitigate the risk of overfitting brought by the absence of anomalous representations,
synthesis-based methods introduce discriminative information through anomaly synthesis based on the two above frameworks.
Generally, the anomaly synthesis strategies can be divided into two levels: image level and feature level.
As depicted in Fig.~\ref{fig:category}(a),
previous works \cite{yang2020anomaly,zavrtanik2021draem,song2022anomaly,yao2022feature,jiang2022masked} follow a common paradigm of
image-level anomaly synthesis by creating binary masks with random positions and shapes, and filling them with various textures from other datasets.
However, they are constrained by the quality of synthetic anomalies and the requirement of auxiliary textures.
Therefore, recent works \cite{you2022unified,liu2023simplenet,lee2024continuous} synthesize feature-level anomalies
by adding Gaussian noise to normal features, as depicted in Fig.~\ref{fig:category}(b).
Since this synthesis strategy is straightforward and the synthetic anomalies do not require duplicate feature extraction,
it can generate and leverage diverse anomalies more efficiently.
Nevertheless, the constant variance and random directions of Gaussian noise across all dimensions
may fail to capture useful anomalous information, leading to considerable redundancy and suboptimal decision boundaries.
Moreover, the natural properties of synthetic anomalies still need to be predefined.
As a result, the performance of existing methods is significantly limited.

\begin{figure*}[t]
    \centering
    \includegraphics[width=0.99\linewidth]{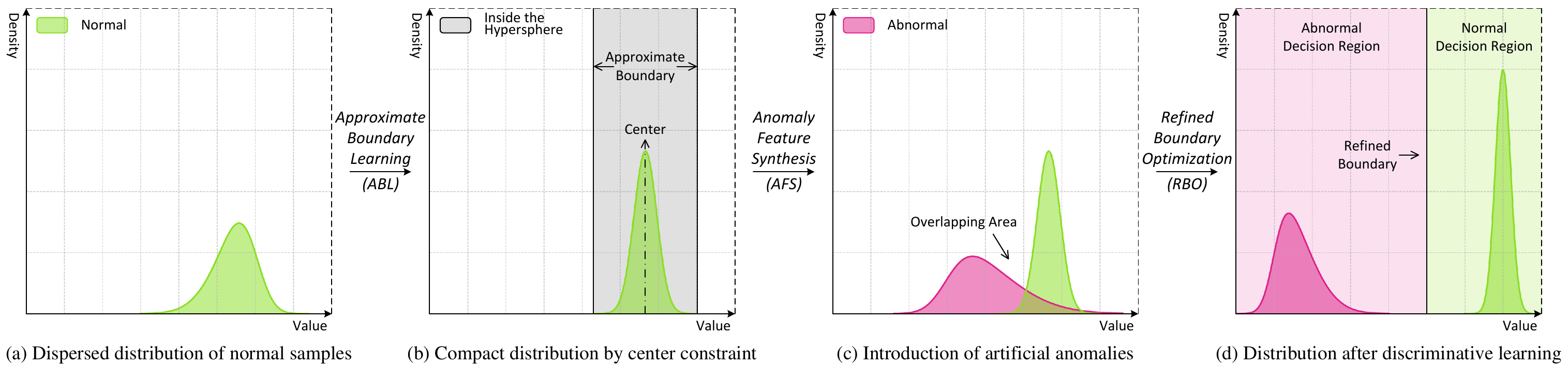}
    \caption{Conceptual illustration of our proposed PBAS.
    (a) The initial distribution of normal samples in the feature space is relatively dispersed.
    (b) Through the learning of ABL, normal features are projected into a compact hypersphere.
    (c) Through the synthesis of AFS, artificial anomalies are generated from normal features by the hypersphere distribution.
    (d) Through the optimization of RBO, the decision boundary is further refined by the discriminative network.
    }
    \label{fig:concept}
\end{figure*}

To address the aforementioned issues,
we propose a novel Progressive Boundary-guided Anomaly Synthesis (PBAS) strategy,
which can directionally synthesize crucial feature-level anomalies without any predefined properties.
It consists of three core components: Approximate Boundary Learning (ABL),
Anomaly Feature Synthesis (AFS), and Refined Boundary Optimization (RBO).
As illustrated in Fig.~\ref{fig:concept}(a-b),
since the distribution of normal samples is relatively dispersed in the feature space,
ABL learns the compact distribution of pretrained normal features by center constraint.
To find a more representative center that captures the intra-class diversity of normal patterns,
we improve the center initialization through the feature alignment with iterative updates.
The hypersphere decision boundary learned by ABL is considered as an approximate boundary,
which is then used to guide the anomaly synthesis in AFS.
Without relying on any auxiliary datasets or pixel-level annotations, AFS synthesizes feature-level anomalies at a flexible length
along the ray direction from the hypersphere center to normal features, as depicted in Fig.~\ref{fig:category}(c).
The ray direction ensures that crucial anomalies are synthesized outside the normal feature space along the fastest path.
The flexible length is self-adaptively determined by the average distance between normal features and center during training.
This mechanism enables the anomaly synthesis in a more controllable manner,
mitigating the overfitting that may arise from mapping all normal samples to one point.
However, the distribution of normal features is more concentrated in specific directions emanating from the center,
while it is more dispersed in other directions.
Since the hypersphere boundary is so loose that it may contain real anomalies,
RBO further optimizes the decision boundary through the binary classification of artificial anomalies and normal features.
As illustrated in Fig.~\ref{fig:concept}(c-d), the refined boundary obtained by discriminative learning can
effectively separate normal features from artificial anomalies without any overlapping area.

In this way, PBAS can directionally synthesize diverse and crucial anomalies guided by the progressive decision boundary,
thereby improving the performance of anomaly detection and localization.
In summary, the main contributions of our proposed PBAS are as follows:
\begin{itemize}
    \item 
    We propose ABL to learn a compact distribution of normal features through iterative feature alignment,
    capturing intra-class diversity and establishing an approximate boundary for guiding anomaly synthesis.
    \item
    We design a novel AFS strategy,
    which synthesizes feature-level anomalies along a ray direction from the hypersphere center.
    AFS also uses a self-adaptive length to control synthesis,
    effectively reducing overlap with normal features and improving efficiency.
    \item
    We introduce RBO to refine the hypersphere boundary through binary classification of synthetic anomalies and normal features,
    achieving state-of-the-art (SOTA) performance in anomaly detection and localization.
\end{itemize}

The remainder of the paper is organized as follows:
Section~\ref{sec:related} introduces related work on unsupervised anomaly detection;
Section~\ref{sec:method} outlines the details of our proposed method;
Section~\ref{sec:exper} presents the experimental results and discussions;
Section~\ref{sec:conclu} provides the conclusions.
\section{Related work}
\label{sec:related}

In this section, we review the typical and SOTA methods for unsupervised anomaly detection.
The existing methods can be divided into three categories.
In brief, reconstruction-based methods aim to reconstruct the input data from the latent space,
while embedding-based methods aim to learn the embedding space for the input data \cite{xie2024iad}.
Based on these two frameworks, 
synthesis-based methods aim to synthesize artificial anomalies to assist in the training of models.

\subsection{Reconstruction-based Methods}
\label{sec:_recon_method}

Reconstruction-based methods have been widely explored in anomaly detection.
It is assumed that the model can properly reconstruct normal samples, while it fails to do so for anomalies.
The key to these methods is to detect anomalies by analyzing the residual images before and after reconstruction.
The Autoencoder (AE) \cite{hinton2006reducing} is a classic model widely used for image reconstruction tasks.
Several studies \cite{zong2018deep,gong2019memorizing,xing2023visual} employ feature reduction and sparse representation methods
to compress the latent representations of AEs, achieving more stable reconstruction results.
However, since AEs have never trained on real anomalies, the assumption that anomalous regions will not be accurately reconstructed does not always hold.
To address this issue, several methods \cite{zavrtanik2021reconstruction,pirnay2022inpainting,luo2024ami} randomly remove some patches,
and reconstruct the missing information through inpainting.
Given the superior reconstruction capability of generative models, 
Generative Adversarial Networks (GANs) \cite{goodfellow2014generative} and Variational Autoencoders (VAEs) \cite{kingma2014auto}
are commonly used frameworks for reconstruction-based models.
Several works \cite{schlegl2017unsuper,akcay2018ganomaly,hou2021divide} train GANs through adversarial learning,
using the reconstruction error of the generator to detect anomalies.
Similarly, several papers \cite{bergmann2019improving,chen2020unsupervised,dehaene2020anomaly} train VAEs to model the distribution of normal samples in latent space,
using the reconstruction error of decoder to detect anomalies.
However, these methods heavily rely on the quality of reconstructed images,
which faces challenges in difference analysis.

\subsection{Embedding-based Methods}
\label{sec:_embed_method}

Embedding-based methods have demonstrated outstanding performance in anomaly detection and localization,
becoming increasingly prevalent in recent years.
These methods utilize pretrained networks to extract features and compress normal features into a compact space.
Consequently, anomaly features are distinctly segregated from normal clusters within the feature space.
The key to these methods is to detect anomalies by calculating the distance between the representations of test images and normal clusters.
First, Knowledge Distillation (KD) methods exploit the disparity in anomaly detection capability between teacher and student networks.
To learn a more robust and generalizable representation, MKD \cite{salehi2021multiresolution} and GLCF \cite{yao2024learning} employ
feature distillation at various layers from the teacher network to the student network.
Since structurally similar teacher-student networks can hinder the diversity of anomalous representations,
RD4AD \cite{deng2022anomaly} and ADPS \cite{xing2024adps} adopt an asymmetrical ``reverse distillation'' paradigm.
Specifically, ADPS concatenates spatial attention-weighted \cite{li2021ctnet} teacher features with
decoded student features to achieve more precise anomaly segmentation.
To further enhance inference speed, EfficientAD \cite{batzner2023efficientad} refines the architecture of teacher network with efficient feature extractor.
Second, memory bank methods store representative normal features and detect anomalies through the distance between test samples and memorized samples.
PaDiM \cite{defard2021padim} memorizes the multivariate Gaussian distributions of normal patch embeddings
and calculates the anomaly score by Mahalanobis distance.
However, PaDiM stores a specific distribution for each patch position.
As PatchCore \cite{roth2022towards} generally stores patches from all positions through the greedy coreset mechanism,
it reduces reliance on image alignment.
PNI \cite{bae2023pni} further integrates position and neighborhood information into the inference stage of PatchCore.
Third, normalizing flow methods \cite{rudolph2022fully,yao2023dual,zhou2024msflow} aim to transform the distribution of normal samples
into a standard Gaussian distribution, resulting in anomalies exhibiting low likelihood.
Finally, one-class classification methods constrain the implicit classification boundaries of normal features by designing loss functions.
A primary paradigm of these methods is to construct decision boundaries that encompass typical normal samples and detect anomalies
by measuring the distance between test samples and the normal center.
SMCC \cite{liu2023unsupervised} utilizes a Gaussian mixture model to obtain cluster centers.
NoCoAD \cite{gao2023exploring} improves the optimization objective by leveraging well-designed norm
based on Deep Support Vector Data Description (Deep-SVDD) \cite{ruff2018deep}.
To address the singular training objective of Deep-SVDD,
CFA \cite{lee2022cfa} searches hard negative features from normal samples to perform contrastive supervision.
Despite achieving superior performance, these models have only been exposed to normal samples.

\subsection{Synthesis-based Methods}
\label{sec:_synth_method}

Synthesis-based methods treat anomaly synthesis as data augmentation for normal samples,
integrating this strategy into reconstruction-based and embedding-based frameworks.
Anomaly detection models trained solely on normal samples lack the ability to learn anomalous distribution.
Leveraging the reconstruction-based framework,
most methods synthesize anomalies at the image level to assist training in a self-supervised manner.
Some works
\cite{yang2020anomaly,zavrtanik2021draem,song2022anomaly,yao2022feature,jiang2022masked,wang2023dual,zhao2023omnial,xing2023normal,wang2024produce}
follow a common paradigm of synthesizing anomalies by creating binary masks using Perlin noise \cite{perlin1985image} and
filling them onto normal images with various textures from auxiliary datasets.
To avoid introducing auxiliary images, Yan et al. \cite{yan2021unsupervised} directly adds random noise to the entire images to simulate anomalies. 
However, synthesizing anomalies at the image level requires manually predefining the visual properties of anomalies.
Without the need for explicit visual guidance, DSR \cite{zavrtanik2022dsr} and IGD \cite{chen2022deep} synthesize anomalies
in feature space through vector replacement and vector weighting.
Leveraging the embedding-based framework,
several studies \cite{yang2023memseg,zhang2023destseg,cai2023discrepancy} using image-level anomaly synthesis
follow the same common paradigm described above.
CutPaste \cite{li2021cutpaste} and Pull\&Push \cite{zhou2022pull}
employ a direct approach by cutting normal regions and pasting them at random positions.
To improve the unnaturalness of direct overlay, NSA \cite{schluter2022natural} uses Poisson image editing to seamlessly blend various images.
AnomalyDiffusion \cite{hu2024anomalydiffusion} and RealNet \cite{zhang2024realnet} utilize
the diffusion model \cite{nichol2021improved} to synthesize anomalies that are more realistic.
Unlike the aforementioned image blending, CDO \cite{cao2023collaborative} and RD++ \cite{tien2023revisiting} add random noise within rectangular masks.
Nevertheless, more realistic image-level anomaly synthesis requires significant computational resources, which greatly affects training speed.
In contrast, feature-level anomaly synthesis is more efficient because it is straightforward and does not require repeated feature extraction.
UniAD \cite{you2022unified} and SimpleNet \cite{liu2023simplenet} synthesize feature-level anomalies by adding Gaussian noise to the normal features.
To enhance the detection of weak anomalies,
GLASS \cite{chen2024unified} further refines the noise distribution by adversarial learning.
However, these methods only cover a fixed range of anomalies in random directions,
which may fail to capture useful information and lead to significant redundancy.  
In contrast, our proposed method, PBAS,
introduces a novel feature-level anomaly synthesis strategy that efficiently generates anomalies with directional guidance and self-adaptive lengths,
offering more control and effectiveness than existing methods that rely on predefined textures or simple noise.
\section{Proposed Method}
\label{sec:method}

\begin{figure*}[t]
  \centering
  \includegraphics[width=0.99\linewidth]{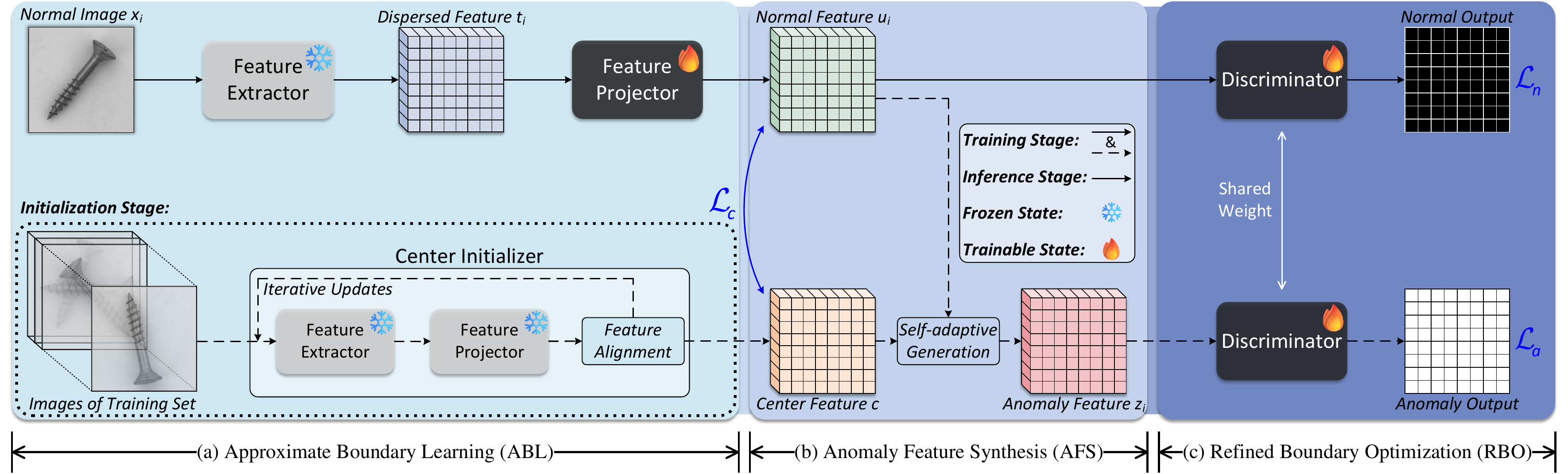}
  \caption{Schematic of the proposed PBAS. 
  (a) Approximate Boundary Learning (ABL) learns an approximate boundary of normal features through the improved center constraint.
  (b) Anomaly Feature Synthesis (AFS) synthesizes anomaly features based on the hypersphere distribution of the ABL output.
  (c) Refined Boundary Optimization (RBO) further refines the boundary through discriminative learning of the AFS output.
  The training stage is depicted with solid and dashed arrows, while the inference stage is indicated by solid arrows.
  }
  \label{fig:schematic}
\end{figure*}

\subsection{Overview}
\label{sec:_overview}

The overall architecture of the proposed PBAS is shown in Fig.~\ref{fig:schematic}.
During the training stage, PBAS consists of three core components: ABL, AFS, and RBO.
The ABL module (Section~\ref{sec:_approximate}) employs a feature extractor ${{E}_{\phi}}$, a feature projector ${{P}_{\theta}}$,
and a center initializer ${{C}_{\phi,\theta}}$ to learn an approximate boundary of normal images.
The backbone of ${{E}_{\phi}}$ (Section~\ref{sec:_extract}) is pretrained on ImageNet and kept frozen.
${{P}_{\theta}}$ within ${{C}_{\phi,\theta}}$ is only frozen at the initialization stage, while it becomes trainable afterward. 
The AFS module (Section~\ref{sec:_synth}) takes the normal and center feature outputs from ABL as inputs and is designed to
synthesize anomaly features self-adaptively, guided by the hypersphere distribution of normal features.
The RBO module (Section~\ref{sec:_refined}) takes the normal and anomaly feature outputs from AFS as inputs and
utilizes a pair of discriminators ${{D}_{\psi}}$ to refine the boundary.
These two ${{D}_{\psi}}$ are trainable and share the same weights.
PBAS is trained in a multi-task learning manner with three loss functions from ABL and RBO.
At the inference stage (Section~\ref{sec:_infer}), only the feature extractor ${{E}_{\phi}}$ and the feature projector ${{P}_{\theta}}$ from ABL,
along with the discriminator ${{D}_{\psi}}$ from RBO, are used.

\subsection{Feature Extraction with Pretrained Model}
\label{sec:_extract}

The pretrained networks can be utilized to extract features from different scales and channels.
In this paper, we employ the ResNet-like backbone $\phi$ pretrained on ImageNet with frozen parameters.
The training set \(X_\text{train}\) for anomaly detection tasks only contains normal images. 
During the training phase, normal images $x_i$ are first fed into the backbone of the feature extractor ${{E}_{\phi}}$
to obtain pretrained features \mbox{\(\phi_{i,j} = \phi_j(x_i) \in \mathbb{R}^{H_j \times W_j \times C_j}\)} at different hierarchy levels $j$.
The feature point at location \mbox{\((h, w)\)} is denoted by \mbox{\(\phi_{i,j}^{h,w} \in \mathbb{R}^{C_j}\)}.
The relationship of this vector to the feature map \(\phi_{i,j}\) is:
\begin{equation}
{\phi _{i,j}} = \left\{ {\left. {\phi _{i,j}^{h,w}} \right|h \in [1, \ldots ,{H_j}],w \in [1, \ldots ,{W_j}]} \right\}
\label{eq:point_to_map}
\end{equation}

To increase the receptive field size and robustness to small spatial deviations,
pretrained features are then aggregated through adaptive pooling \cite{roth2022towards}.
The location set of neighborhood vectors associated with \(\phi_{i,j}^{h,w}\) is:
\begin{align}
&N_p^{h,w} = \left\{ (a,b) \mid a \in \left[ h - \left\lfloor \frac{p}{2} \right\rfloor, \ldots,
h + \left\lfloor \frac{p}{2} \right\rfloor \right], \right. \nonumber \\
&\phantom{N_p^{h,w} = \left\{ (a,b) \mid \right.} \left. b \in \left[ w - \left\lfloor \frac{p}{2} \right\rfloor, \ldots,
w + \left\lfloor \frac{p}{2} \right\rfloor \right] \right\}
\label{eq:neighbour_location}
\end{align}

\noindent
where $p$ denotes the neighborhood size. Hence, the neighborhood aggregation vectors \(s_{i,j}^{h,w}\) can be expressed as:
\begin{equation}
s_{i,j}^{h,w} = {f_{\text{agg}}}\left( {\left\{ {\left. {\phi _{i,j}^{a,b}} \right|(a,b) \in N_p^{h,w}} \right\}} \right)
\label{eq:locally_aware}
\end{equation}

\noindent
where $f_{\text{agg}}$ represents adaptive pooling that integrates the local feature patch into a single feature point.

Using multilevel concatenation $f_{\text{concat}}$ to capture low-level and high-level features,
we obtain the dispersed feature:
\begin{equation}
{t_i} = {E_\phi}({x_i}) = {f_{\text{concat}}}\left( {\left\{ {\left.
{f_{\text{resize}}^{{H_m},{W_m}}\left( {{s_{i,j}}} \right)} \right|j \in J} \right\}} \right)
\label{eq:final_feature}
\end{equation}

\noindent
where $J$ denotes the set of selected hierarchy levels.
Each feature map $s_{i,j}$ is upsampled to the maximum resolution \mbox{$({H_m},{W_m})$} of the lowest hierarchy level using $f_{\text{resize}}$. 

\subsection{Approximate Boundary Learning by Center Constraint}
\label{sec:_approximate}

The features obtained by the feature extractor ${{E}_{\phi}}$ already contain useful information for anomaly detection \cite{xu2022discriminative}. 
However, the distribution of these features is highly dispersed.
Traditional SVDD methods using simple averaging for center estimation fail to capture intra-class variations.
To capture intra-class diversity and establish a compact boundary,
we propose a novel center constraint method called Approximate Boundary Learning (ABL),
which improves the center initialization by iterative feature alignment, as shown in Fig.~\ref{fig:schematic}(a).

\begin{figure*}[t]
  \centering
  \includegraphics[width=0.99\linewidth]{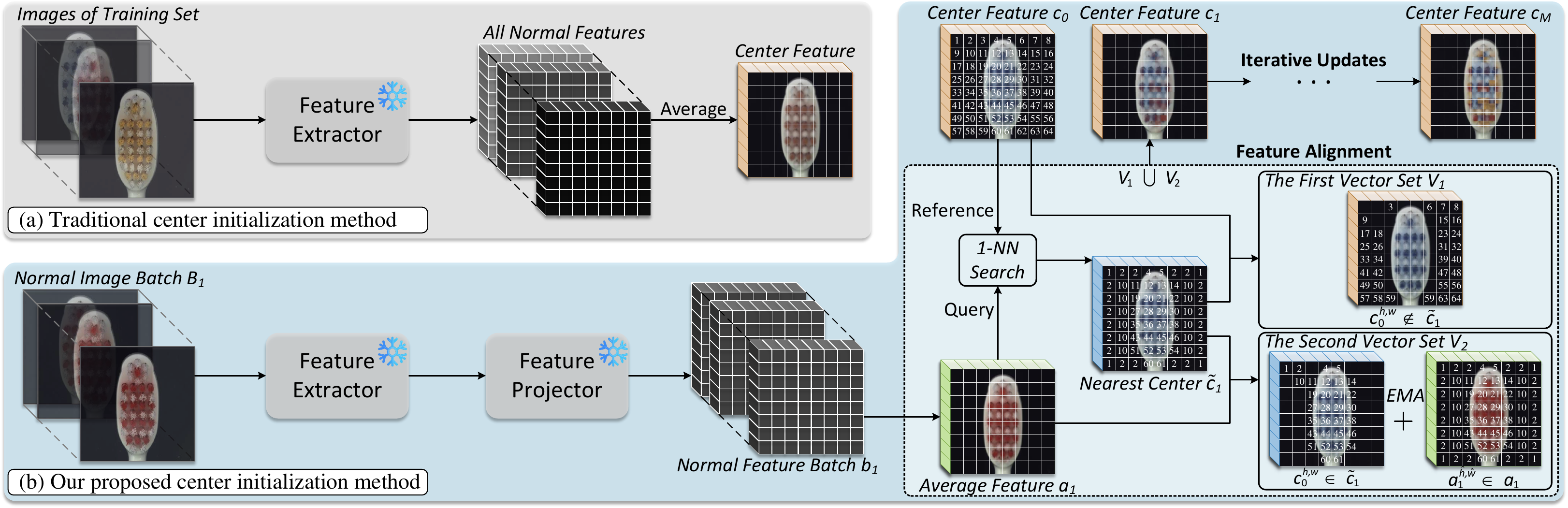}
  \caption{Comparison of center initialization methods. (a) The traditional method obtains center feature by
  the average feature of entire training set through a single feature extractor.
  (b) Our method obtains center feature by feature alignment with iterative updates through a pair of feature extractor and projector.}
  \label{fig:center}
\end{figure*}

\textbf{Center Initialization}. Similar to \cite{ruff2018deep}, ABL aims to project features into another space within
the smallest possible hypersphere that encompasses the majority of normal features.
To begin with, it is essential to determine the center feature before boundary learning.
By utilizing a single feature extractor ${E_\phi}$, the traditional center initialization method
depicted in Fig.~\ref{fig:center}(a) computes the center feature as:
\begin{equation}
\bar c = \frac{1}{N}\sum\limits_{i = 1}^N {{E_\phi }({x_i})}
\label{eq:traditional}
\end{equation}

\noindent
where $N$ is the total number of training images.
Due to the intra-class variations of industrial datasets,
the center feature cannot be accurately computed in this direct way. 

\begin{algorithm}[t]
  \caption{Framework of Approximate Boundary Learning}
  \begin{algorithmic}[1]
  \State \textbf{Input:} training set \(X_\text{train}\), normal batch $B_i$, number of batches $M$,
  normal image $x_i$, distance metric $\mathcal{D}$, and smoothing factor $\beta$ for EMA
  \State \textbf{Output:} normal features \(u_i\), center feature \(c\)
  \State \textit{\# Center Initialization.}
  \For{\(B_i\) in \(X_\text{train}\)}
      \State \(b_i \leftarrow {{P}_{\theta}}({{E}_{\phi}}(B_i))\), with frozen ${E}_{\phi}$ and frozen ${P}_{\theta}$
      \State \(a_i \leftarrow {\bar b}_i\)
      \If{\(b_i\) is the first batch}
          \State \(c_0 \leftarrow a_0\)
      \Else
          \State \textit{\# 1-NN search for each vector in $a_i$ within $c_{i-1}$}
          \State \({\tilde c_{i}} \leftarrow \left\{ {\left. {\mathop {\arg \min }\limits_{c_{i - 1}^{\hat h,\hat w} \in
          {c_{i - 1}}} {{\mathcal{D}\left( {a_i^{h,w},c_{i - 1}^{\hat h,\hat w}}\right)}}} \right|a_i^{h,w} \in {a_i}} \right\}\)
          \vspace{2pt}
          \State \textit{\# Update $c_{i}$ through two sets of vectors}
          \vspace{3pt}
          \State \({c_i} \leftarrow \left\{ {\left. {c_{i - 1}^{h,w}} \right|c_{i - 1}^{h,w} \notin \tilde c_{i}} \right\}\)
          \vspace{3pt}
          \State \({c_i} \cup \left\{ {\left. {\left( {1 - \beta } \right) c_{i - 1}^{h,w} +
          \beta a_i^{\hat h,\hat w}} \right|c_{i - 1}^{h,w} \in \tilde c_{i}},a_{i}^{\hat h,\hat w} \in a_{i} \right\}\)
      \EndIf
  \EndFor
  \State \(c \leftarrow c_M\)
  \State \textit{\# Boundary Learning.}
  \For{\(x_i\) in \(X_\text{train}\)}
      \State \(u_i \leftarrow {{P}_{\theta}}({{E}_{\phi}}(x_i))\), with frozen ${E}_{\phi}$ and trainable ${P}_{\theta}$
      \State \textit{\# 1-NN search for each vector in $u_i$ within $c$}
      \State \({\tilde c} \leftarrow \left\{ {\left. {\mathop {\arg \min }\limits_{{c^{\hat h,\hat w}} \in c}
      {{\mathcal{D}\left( {u_i^{h,w},c^{\hat h,\hat w}}\right)}}} \right|u_i^{h,w} \in {u_i}} \right\}\)
      \vspace{2pt}
      \State Calculate the loss ${\mathcal{L}_{{\text{c}}}}$ of ABL by Eq.~\ref{eq:l_center}
  \EndFor
  \State \textbf{return} normal features \(u_i\), center feature \(c\)
  \end{algorithmic}
  \label{alg:ABL}
  \end{algorithm}

Fig.~\ref{fig:center}(b) outlines our center initializer ${{C}_{\phi,\theta}}$ by the feature alignment with iterative updates.
To determine a more appropriate center feature, we search for the reference center using query feature vectors and update it batch by batch.
The training set is divided into $M$ batches.
Each feature batch is obtained through a pair of frozen feature extractor and projector.
The frozen feature projector $P_\theta$ is initialized with a normal distribution,
producing a Gaussian filter-like effect on features that stabilizes center initialization.
Concretely, the initial center feature \(c_0\) is first derived directly as the average of the feature batch \(b_0\). 
After obtaining the average feature \(a_1\) of normal feature batch \(b_1\),
the nearest center \(\tilde c_1\) is determined through the 1-Nearest Neighbor (1-NN) search
for each vector in the query feature \(a_1\) within reference center \(c_0\).
Then, the center feature \(c_1\) is obtained from the union of two vector sets through feature alignment:
\begin{itemize}
    \item The first set \(V_1\) includes vectors in \(c_0\) but not in \(\tilde c_1\);
    these vectors of \(c_1\) remain consistent with vectors of \(c_0\).
    \item The second set \(V_2\) includes vectors of \(c_0\) within \(\tilde c_1\);
    these vectors of \(c_1\) are updated using the Exponential Moving Average (EMA) of corresponding vectors from \(\tilde c_1\) and \(a_1\).
\end{itemize}
This iterative updates continues until final center feature $c$ (equal to \(c_M\)) is obtained upon completion of the search through the entire training set.

\textbf{Boundary Learning}. Subsequently, the feature projector ${{P}_{\theta}}$ is used to map the dispersed features $t_{i}$ close to the center feature $c$. 
The normal features are denoted as \mbox{${u_i} = {{P}_{\theta}}({t_i})$} where ${{P}_{\theta}}$ is a fully-connected layer with equal input and output nodes.
  Instead of aligning all features directly to a single center, which could lead to mode collapse,
each vector $u_{i}^{h,w}$ is matched with its nearest center vector $\tilde c^{\hat h,\hat w}$ within the center feature $c$.
The loss function of ABL is:
\begin{equation}
{{\cal L}_{{\text{c}}}} = \frac{1}{N}\frac{1}{{{H_m}{W_m}}}\sum\limits_{i = 1}^N
{\sum\limits_{h,w} {\mathcal{D}\left( {u_i^{h,w},{\tilde c^{\hat h,\hat w}}}\right)} }
\label{eq:l_center}
\end{equation}

\noindent
where \( \mathcal{D}(\cdot, \cdot) \) is a predefined distance metric, specifically the Euclidean distance.
Due to the relatively loose nature of the hypersphere, 
ABL only provides an approximate boundary for the normal features at the current stage, 
as depicted by the light green circle of Fig.~\ref{fig:boundary}.
The entire framework of ABL is detailed in Algorithm~\ref{alg:ABL}.

\subsection{Hypersphere-based Anomaly Feature Synthesis}
\label{sec:_synth}
\begin{figure}[t]
  \centering
  \includegraphics[width=0.99\linewidth]{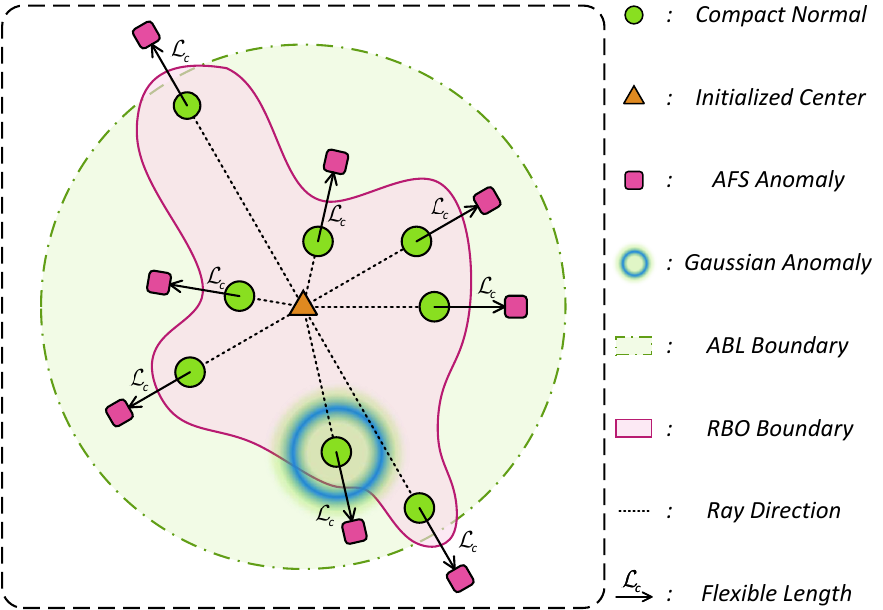}
  \caption{Mechanism of the self-adaptive generation by AFS and progressive boundary guidance by ABL and RBO.}
  \label{fig:boundary}
\end{figure}

Synthesizing anomalies in the feature space has proven to be an effective method for
enhancing anomaly detection tasks \cite{you2022unified,liu2023simplenet}.
As shown in Fig.~\ref{fig:boundary}, these works generate feature-level anomalies by simply adding Gaussian noise without directional constraints,
which may result in Gaussian anomalies still residing within the normal sample space.
As the features become more concentrated during training,
the constant variance of Gaussian noise further increases the likelihood of overlap between synthetic anomalies and the normal feature distribution.
To generate more useful anomalies for boundary optimization, we design the Anomaly Feature Synthesis (AFS) method,
which is based on the hypersphere distribution of normal features,
as shown in Fig.~\ref{fig:schematic}(b).

AFS generates near-in-distribution anomalies derived from the normal features of the original training set.
As shown in Fig.~\ref{fig:boundary},
starting from the normal feature ${u_i}$, the anomaly feature ${z_i}$ is synthesized at a flexible length
along the ray direction from the nearest center feature ${\tilde c}$ to ${u_i}$.
The ray direction ensures that the synthetic anomalies are reliably positioned outside the normal feature space
by following the path that allows for the fastest departure from the distribution of normal features,
thereby minimizing overlap and enabling the most efficient generation.
The self-adaptive generation process is mathematically formulated as:
\begin{equation}
{z_i} = {u_i} + \alpha  \cdot {{\cal L}_{{\text{c}}}} \cdot \frac{{{u_i} - \tilde c}}{{\left\| {{u_i} - \tilde c} \right\|}}
\label{eq:synthesis}
\end{equation}

\noindent
where $\alpha$ is a hyperparameter that controls the range of anomaly synthesis.
As $\alpha$ decreases, it becomes more difficult to distinguish the anomaly feature $z_i$ from the normal feature $u_i$, and vice versa.

In the self-adaptive generation described in Eq.~\ref{eq:synthesis}, the flexible length \(\mathcal{L}_{\text{c}}\) plays a crucial role.
The \(\mathcal{L}_{\text{c}}\) given by Eq.~\ref{eq:l_center} represents the average distance
from nearest center feature \(\tilde c\) to normal feature \(u_i\).
Since all \(u_i\) in a batch share the same \(\mathcal{L}_{\text{c}}\), each anomaly feature \(z_i\) maintains the same distance from \(u_i\).
This ensures that \(z_i\) cannot get too close to \(\tilde{c}\), reducing overlap between normal and anomaly features.
With the iterative training of boundary learning, \(\mathcal{L}_{\text{c}}\) gradually decreases.
Consequently, \(z_i\) is also compressed as the hypersphere shrinks.
This ensures that \(z_i\) cannot get too far from \(u_i\), preventing the anomalies from becoming useless.

\subsection{Refined Boundary Optimization by Discriminative Network}
\label{sec:_refined}

Through the boundary learning of ABL, normal features are constrained within the loose hypersphere boundary,
with the majority of them being centrally clustered near the center feature.
However, the distribution of normal features is more concentrated in specific directions emanating from the center feature,
while it is more dispersed in other directions.
To address the issue of real anomalies potentially occurring within the hypersphere,
the Refined Boundary Optimization (RBO) is introduced to further optimize the boundary based on synthetic anomaly features and normal features,
as shown in Fig.~\ref{fig:schematic}(c).

Inspired by \cite{huang2023enhancing}, RBO employs a discriminator ${D_\psi}$ to enlarge the score disparity between normal and anomaly features,
where ${D_\psi}$ is structured as a three-layer Multi-Layer Perceptron (MLP) with Sigmoid.
The loss ${\cal L}_{{\text{n}}}$ for normal features is given by the
Binary Cross-Entropy (BCE) loss between the normal confidence and ground truth of 0:
\begin{equation}
{{\cal L}_{{\text{n}}}} = \frac{1}{N}\frac{1}{{{H_m}{W_m}}}\sum\limits_{i = 1}^N {\sum\limits_{h,w} {{f_{bce}}\left( {{D_\psi }(u_i^{h,w}),{0}} \right)} }
\label{eq:l_normal}
\end{equation}

The loss ${\cal L}_{{\text{a}}}$ for anomaly features is given by the BCE loss between the anomaly confidence and ground truth of 1:
\begin{equation}
{{\cal L}_{{\text{a}}}} = \frac{1}{N}\frac{1}{{{H_m}{W_m}}}\sum\limits_{i = 1}^N {\sum\limits_{h,w} {{f_{bce}}\left( {{D_\psi }(z_i^{h,w}),{1}} \right)} }
\label{eq:l_anomaly}
\end{equation}

Through the discriminative learning of normal and anomaly features synthesized by AFS,
the boundary obtained by RBO is more refined compared to the hypersphere boundary obtained by ABL,
as depicted by the light pink region in Fig.~\ref{fig:boundary}.
By leveraging the joint training with near-in-distribution anomalies,
RBO effectively mitigates the risk of model collapse in ABL,
where optimizing the feature projector $P_\theta$ with a single center constraint loss could potentially map all normal features to one point.
According to Eqs.~\ref{eq:l_center}, \ref{eq:l_normal}, and \ref{eq:l_anomaly}, the overall training objective of PBAS is:
\begin{align}
\mathcal{J}(\theta, \psi) = & \mathop {\min }\limits_\theta  \left( {{{\cal L}_{{\text{c}}}} + \gamma {{\left\| \theta  \right\|}^2}} \right) \nonumber \\
& + \min_{\theta, \psi} \left(\mathcal{L}_{\text{n}} + \mathcal{L}_{\text{a}} +
\delta ({{\left\| \theta  \right\|}^2} + {{\left\| \psi  \right\|}^2})\right)
\label{eq:l_overall}
\end{align}

\noindent
where $\gamma$ and $\delta$ represent the regularization coefficients for enhancing the generalization ability of ABL and RBO.

During the iterative training of PBAS, normal features contract towards the center with the boundary evolving from approximate to refined.
In summary, our proposed anomaly synthesis strategy is guided by the progressive boundary refinement to enhance anomaly detection.

\subsection{Anomaly Scoring at Inference Stage}
\label{sec:_infer}

As depicted in Fig.~\ref{fig:schematic}, the inference stage is represented by the solid arrows without AFS. 
The test set \(X_\text{test}\) contains both normal and abnormal images.
Concretely, the test image \(x_i\) is processed by the feature extractor \(E_{\phi}\) and
the feature projector \(P_{\theta}\) in ABL to obtain the test feature \(u_i\).
Then, the discriminator ${{D}_{\psi}}$ in RBO directly outputs the confidence scores.
Since the center feature provided by ABL is not required during inference,
PBAS allows for a streamlined and efficient inference process.

\textbf{Anomaly Detection} refers to the task of classifying test images as either normal or anomalous,
essentially a classification task at the image level.
The image-level anomaly score ${\cal S}_{{\text{AD}}}$ of test image \(x_i\) is calculated by the maximum value of all vectors in test feature ${u_i}$:
\begin{equation}
  {{\cal S}_{{\text{AD}}}} = \mathop {\max }\limits_{u_i^{h,w} \in {u_i}} {D_\psi }(u_i^{h,w})
  \label{eq:score_ad}
\end{equation}

\noindent
where ``AD'' denotes Anomaly Detection.

\textbf{Anomaly Localization} refers to the task of identifying the specific locations of anomalies within test images,
essentially a segmentation task at the pixel level.
First, bilinear interpolation is employed to upsample the confidence scores from feature dimensions \mbox{\(({H_m},{W_m})\)} to
image dimensions \mbox{\(({H_0},{W_0})\)} using \(f_{\text{resize}}\).
Second, Gaussian smoothing is applied to reduce noise using \(f_{\text{smooth}}\).
Finally, the pixel-level anomaly score \({\cal S}_{\text{AL}}\) of the test image \(x_i\) is calculated by:
\begin{equation}
  {{\cal S}_{{\text{AL}}}} = {f_{\text{smooth}}}(f_{\text{resize}}^{{H_0},{W_0}}({D_\psi }({u_i})))
  \label{eq:score_al}
\end{equation}

\noindent
where ``AL'' denotes Anomaly Localization.   
\section{Experiment and Analysis}
\label{sec:exper}

\begin{table*}[t]\scriptsize 
  \centering
  \caption{Performance comparison of different methods on each category of MVTec AD, as measured by I-AUROC\%.
  The best results for each category are highlighted in bold, and the second-best results are underlined.
  }
    \begin{tabularx}{0.99\textwidth}{X|*{3}{>{\centering\arraybackslash}X}|*{8}{>{\centering\arraybackslash}X}|>{\centering\arraybackslash}X}
    \hline
    \multirow{3}{*}{Category} & \multicolumn{3}{c|}{Embedding-based methods} & \multicolumn{9}{c}{Synthesis-based methods} \\
    \cline{2-13}         & NoCoAD & CFA   & RD4AD & Pull\&Push & CutPaste & DRAEM & DSR   & DBPI  & DeSTSeg & RD++  & SimpleNet & \multirow{2}{*}{PBAS} \\
      & \cite{gao2023exploring}   & \cite{lee2022cfa}   & \cite{deng2022anomaly}   & \cite{zhou2022pull}   & \cite{li2021cutpaste}   & \cite{zavrtanik2021draem}  
      & \cite{zavrtanik2022dsr}   & \cite{wang2023dual}   & \cite{zhang2023destseg}  & \cite{tien2023revisiting}   & \cite{liu2023simplenet}  &  \\
    \hline
    Carpet & 99.1  & 94.7  & 98.7  & 95.9  & 93.9  & 96.3  & 99.6  & 99.0  & 99.3  & \textbf{100} & \underline{99.7}  & \textbf{100} \\
    Grid  & 98.8  & 99.7  & \textbf{100} & \underline{99.9}  & \textbf{100} & \textbf{100} & \textbf{100} & 98.4  & 98.6  & \textbf{100} & \underline{99.9}  & \textbf{100} \\
    Leather & \textbf{100} & 98.6  & \textbf{100} & 63.6  & \textbf{100} & \textbf{100} & \underline{99.3}  & \textbf{100} & \textbf{100} & \textbf{100} & \textbf{100} & \textbf{100} \\
    Tile  & \textbf{100} & \underline{99.9}  & 99.3  & 99.7  & 94.6  & \textbf{100} & \textbf{100} & \textbf{100} & \textbf{100} & 99.7  & 98.7  & \textbf{100} \\
    Wood  & 99.4  & 97.4  & 99.2  & 99.6  & 99.1  & 99.2  & 94.7  & 99.6  & \underline{99.7}  & 99.3  & 99.5  & \textbf{99.8} \\
    \hline
    Bottle & \textbf{100} & \textbf{100} & \underline{99.9}  & \underline{99.9}  & 98.2  & 96.9  & 99.6  & 99.4  & \textbf{100} & \textbf{100} & \textbf{100} & \textbf{100} \\
    Cable & 95.2  & 99.4  & 97.0  & 98.4  & 81.2  & 93.4  & 95.3  & 95.9  & 97.3  & 99.3  & \textbf{100} & \underline{99.8} \\
    Capsule & 94.9  & \textbf{99.9} & 98.2  & \underline{99.8}  & 98.2  & 96.1  & 98.3  & 97.1  & 97.8  & 99.0  & 97.8  & 99.7 \\
    Hazelnut & \textbf{100} & 99.6  & \textbf{100} & 99.5  & 98.3  & \textbf{100} & 97.7  & 98.8  & \textbf{100} & \textbf{100} & \underline{99.8}  & \textbf{100} \\
    Metal nut & 99.7  & 96.5  & \textbf{100} & 86.9  & \underline{99.9}  & 99.4  & 99.1  & \textbf{100} & 99.2  & \textbf{100} & \textbf{100} & \underline{99.9} \\
    Pill  & 96.6  & 98.5  & 96.8  & \textbf{99.7} & 94.9  & 96.8  & \underline{98.9}  & 95.2  & 97.1  & 98.4  & 98.6  & 98.7 \\
    Screw & 95.3  & 98.3  & 97.8  & 85.3  & 88.7  & \textbf{99.2} & 95.9  & 98.1  & 88.3  & \underline{98.9}  & 98.7  & 98.5 \\
    Toothbrush & 97.2  & \textbf{100} & \underline{99.7}  & 94.0  & 99.4  & \underline{99.7}  & \textbf{100} & \textbf{100} & \textbf{100} & \textbf{100} & \textbf{100} & \textbf{100} \\
    Transistor & 93.0  & 99.4  & 96.8  & \textbf{100} & 96.1  & 94.2  & 96.3  & 99.7  & \underline{99.8}  & 98.5  & \textbf{100} & \textbf{100} \\
    Zipper & 97.2  & 88.8  & 98.0  & 99.6  & \underline{99.9}  & 99.7  & 98.5  & 93.8  & \textbf{100} & 98.6  & \underline{99.9}  & \textbf{100} \\
    \hline
    Average & 97.7  & 98.1  & 98.8  & 94.8  & 96.2  & 98.1  & 98.2  & 98.3  & 98.5  & 99.4  & \underline{99.5}  & \textbf{99.8} \\
    \hline
    \end{tabularx}%
  \label{tab:mvtec_compare_a}%
\end{table*}%

\begin{table*}[t]\scriptsize 
  \centering
  \caption{Performance comparison of different methods on each category of MVTec AD, as measured by P-AUROC\%/P-PRO\%.
  The best results for each category are highlighted in bold, and the second-best results are underlined.}
    \begin{tabularx}{0.99\textwidth}{X|*{3}{>{\centering\arraybackslash}X}|*{8}{>{\centering\arraybackslash}X}|>{\centering\arraybackslash}X}
    \hline
    \multirow{3}{*}{Category} & \multicolumn{3}{c|}{Embedding-based methods} & \multicolumn{9}{c}{Synthesis-based methods} \\
    \cline{2-13}         & NoCoAD & CFA   & RD4AD & Pull\&Push & CutPaste & DRAEM & DSR   & DBPI  & DeSTSeg & RD++  & SimpleNet & \multirow{2}{*}{PBAS} \\
      & \cite{gao2023exploring}   & \cite{lee2022cfa}   & \cite{deng2022anomaly}   & \cite{zhou2022pull}   & \cite{li2021cutpaste}   & \cite{zavrtanik2021draem}  
      & \cite{zavrtanik2022dsr}   & \cite{wang2023dual}   & \cite{zhang2023destseg}  & \cite{tien2023revisiting}   & \cite{liu2023simplenet}  &  \\
    \hline
    Carpet & 99.0/96.7 & 98.5/86.8 & 98.9/97.0 & \textbf{99.5/98.3} & 98.3/~~~-~~~    & 96.2/92.0 & 96.0/94.0 & 98.4/91.0 & 94.7/87.8 & \underline{99.2}/\underline{97.7} & 98.4/85.9 & 99.0/95.8 \\
    Grid  & 98.0/94.4 & 97.8/94.8 & 99.3/97.6 & \underline{99.4}/97.7 & 97.5/~~~-~~~ & \textbf{99.6}/97.8 & \textbf{99.6}/\textbf{99.0} & 95.6/93.0 & 97.9/94.2 & 99.3/97.7 & 98.5/94.3 & 99.0/\underline{97.9} \\
    Leather & 99.4/97.8 & 98.3/93.3 & 99.4/\underline{99.1} & \textbf{99.7}/98.7 & 99.5/~~~-~~~ & 98.9/96.8 & \underline{99.5}/97.9 & 97.1/94.7 & \textbf{99.7}/99.0 & \underline{99.5}/\textbf{99.2} & 99.2/96.4 & 99.4/98.5 \\
    Tile  & 97.2/91.8 & 98.2/93.5 & 95.7/90.6 & 96.8/90.3 & 90.5/~~~-~~~ & \textbf{99.5}/97.4 & 98.6/97.0 & \underline{99.3}/\underline{99.1} & \textbf{99.5}/98.4 & 96.6/92.4 & 97.7/89.5 & 97.6/\textbf{99.3} \\
    Wood  & 95.7/92.1 & \underline{96.7}/92.8 & 95.4/90.9 & 95.2/\underline{93.2} & 95.5/~~~-~~~ & \textbf{97.2}/92.8 & 91.5/88.6 & 95.1/90.2 & 95.4/88.0 & 95.8/\textbf{93.3} & 94.4/83.8 & 95.8/92.2 \\
    \hline
    Bottle & 98.5/95.0 & 98.4/93.1 & \underline{98.8}/96.8 & 98.7/95.6 & 97.6/~~~-~~~ & \textbf{99.3}/96.4 & \underline{98.8}/95.1 & 98.0/93.8 & 99.3/96.8 & \underline{98.8}/\underline{97.0} & 98.0/88.2 & 98.7/\textbf{97.1} \\
    Cable & 97.3/90.8 & \textbf{98.5}/91.7 & 97.0/90.6 & 95.7/87.5 & 90.0/~~~-~~~ & 95.4/75.4 & 97.7/88.2 & 93.6/90.0 & 96.9/89.9 & \underline{98.4}/\underline{93.9} & 97.5/90.1 & \textbf{98.5}/\textbf{98.1} \\
    Capsule & \underline{98.9}/94.8 & 93.4/76.2 & 98.6/95.9 & 97.8/89.9 & 97.4/~~~-~~~ & 94.0/90.4 & 91.0/83.6 & 98.6/95.2 & \textbf{99.2}/\underline{97.3} & 98.8/96.4 & \underline{98.9}/91.4 & \textbf{99.2}/\textbf{98.0} \\
    Hazelnut & 98.9/\underline{96.8} & 93.2/86.1 & 99.0/95.5 & 98.6/96.1 & 97.3/~~~-~~~ & \textbf{99.5}/97.5 & 99.1/90.5 & 98.7/96.1 & \underline{99.4}/\textbf{98.1} & 99.2/96.3 & 98.1/77.3 & 99.1/95.6 \\
    Metal nut & \textbf{99.3}/96.1 & 95.7/84.7 & 97.3/92.4 & 97.8/93.2 & 93.1/~~~-~~~ & 98.7/93.2 & 94.1/91.2 & 98.2/97.3 & \textbf{99.3}/\textbf{97.7} & 98.1/93.0 & 98.8/86.0 & \underline{99.1}/\underline{97.4} \\
    Pill  & \underline{99.1}/\underline{97.6} & 98.5/92.6 & 98.2/96.4 & 98.6/94.9 & 95.7/~~~-~~~ & 97.6/88.1 & 94.2/93.4 & 97.4/96.5 & 98.8/96.4 & 98.3/97.0 & 98.6/93.7 & \textbf{99.2}/\textbf{98.0} \\
    Screw & 98.6/95.5 & 97.7/92.8 & \underline{99.6}/\underline{97.8} & 96.9/85.6 & 96.7/~~~-~~~ & \textbf{99.7}/97.0 & 98.1/88.3 & 99.1/92.9 & 96.7/92.2 & \textbf{99.7}/\textbf{98.6} & 99.2/94.8 & 99.2/96.9 \\
    Toothbrush & 99.0/91.7 & 99.1/96.5 & 99.1/94.5 & 99.1/91.8 & 98.1/~~~-~~~ & 98.1/89.9 & \underline{99.5}/94.5 & 99.0/89.8 & \textbf{99.6}/\underline{98.3} & 99.1/94.2 & 98.5/92.0 & 98.9/\textbf{99.6} \\
    Transistor & 92.7/84.0 & 97.5/89.6 & 93.0/79.4 & \textbf{99.2}/\textbf{97.6} & 93.0/~~~-~~~ & 90.0/81.0 & 80.3/80.2 & 95.6/90.6 & 95.5/91.9 & 94.3/81.8 & 97.0/91.6 & \underline{97.7}/\underline{96.7} \\
    Zipper & 98.7/96.0 & 95.7/83.2 & 98.2/95.5 & 97.9/93.0 & \textbf{99.3}/~~~-~~~ & 98.6/95.4 & 98.4/94.2 & 98.5/86.9 & 98.4/95.0 & 98.8/\underline{96.4} & \underline{98.9}/95.7 & 98.8/\textbf{98.1} \\
    \hline
    Average & 98.0/94.1 & 97.2/89.9 & 97.8/94.0 & 98.1/93.6 & 96.0/~~~-~~~ & 97.5/92.1 & 95.8/91.7 & 97.5/93.1 & 98.0/94.7 & \underline{98.3}/\underline{95.0} & 98.1/90.0 & \textbf{98.6}/\textbf{97.3} \\
    \hline
    \end{tabularx}%
  \label{tab:mvtec_compare_b}%
\end{table*}%

\subsection{Datasets}
\label{sec:_Dataset}

In the experiments, we use three publicly available real-world datasets
renowned for their broad application in the field of industrial anomaly detection.

\textit{1) MVTec AD:}
The MVTec Anomaly Detection \cite{bergmann2019mvtec} dataset is one of the most challenging datasets in the domain. 
This dataset contains 15 high-resolution industrial product categories divided into
texture and object groups with 5354 images, including over 70 types of defects. 
The training set comprises 3629 normal samples, while the test set contains 467 normal samples and 1258 anomalous samples.

\textit{2) VisA:}
The Visual Anomaly \cite{zou2022spot} dataset is one of the largest datasets for industrial anomaly detection,
including 10821 images across 12 categories of colored industrial parts. 
The training set comprises 8659 normal samples, while the test set contains 962 normal samples and 1200 anomalous samples.

\textit{3) MPDD:}
The Metal Parts Defect Detection \cite{jezek2021deep} dataset contains 1346 images of metal parts under varied camera conditions across 6 categories.
The training set comprises 888 normal samples, while the test set contains 176 normal samples and 282 anomalous samples.

\subsection{Implementation Details}
\label{sec:_implement}

\textit{1) Experimental Settings:}
All experiments are conducted using an Intel\textregistered~Xeon\textregistered~Gold 6226R CPU @2.90GHz and an NVIDIA GeForce RTX 3090 GPU.
During the training stage, input images for all methods are resized and center-cropped to the resolution of 256\(\times\)256.
The Adam optimizer is employed to train the feature projector \(P_{\theta}\) and the discriminator \(D_{\psi}\),
with learning rates of \(10^{-4}\) and \(2\times10^{-4}\), respectively.
The training process consists of 400 epochs, with the batch size of 8.
For example, training the ``Carpet'' class from the MVTec AD dataset takes approximately 2.5 hours,
with each image requiring an average training time of 80 ms.
Our proposed PBAS framework comprises three components.
For ABL, we utilize the WideResnet50 \cite{zagoruyko2016wide} pretrained on ImageNet as the backbone for the feature extractor \(E_{\phi}\).
Then, we concatenate the features from hierarchy levels 2 and 3.
The neighborhood patch size \(p\) is set to 3, and the smoothing factor \(\beta\) for EMA is set to 0.1.
For AFS, the anomaly degree \(\alpha\) is set to 0.3.
For RBO, the regularization coefficients \(\gamma\) and \(\delta\) are set to \(10^{-5}\) and \(10^{-2}\), respectively.

\textit{2) Evaluation Metrics:}
To effectively evaluate the discriminative capability of different models at both image and pixel levels,
we employ the Area Under the Receiver Operating Characteristic Curve (AUROC) during the inference stage.
Given its independence from predefined classification thresholds and its robustness against class imbalance,
AUROC is a widely adopted evaluation metric in anomaly detection and localization.
AUROC at the image and pixel levels are denoted as I-AUROC and P-AUROC, respectively.
To evaluate the precision-recall balance in anomaly detection,
this paper follows \cite{xing2024recover,xing2024adps} by using Average Precision (AP),
which is more informative for imbalanced datasets than AUROC.
AP at the image and pixel levels are denoted as I-AP and P-AP, respectively.
For a more comprehensive assessment of the ability to localize anomalies,
we additionally calculate the Per-Region Overlap (PRO) \cite{bergmann2020uninformed} at the pixel level.
PRO is particularly sensitive to smaller-scale anomalies and avoids the risk of AUROC overestimating performance due to an increase in false positives.
PRO at the pixel level is denoted as P-PRO.

\subsection{Comparative Experiments}
\label{sec:_comparative}

To evaluate our proposed PBAS, several typical and SOTA methods are employed in comparative experiments.
NoCoAD \cite{gao2023exploring}, CFA \cite{lee2022cfa}, and RD4AD \cite{deng2022anomaly} employ an embedding-based framework without anomaly synthesis.
In contrast, DRAEM \cite{zavrtanik2021draem}, DSR \cite{zavrtanik2022dsr}, and DBPI \cite{wang2023dual}
employ a reconstruction-based framework integrated with anomaly synthesis.
Similarly, Pull\&Push \cite{zhou2022pull}, CutPaste \cite{li2021cutpaste},
DeSTSeg \cite{zhang2023destseg}, RD++ \cite{tien2023revisiting}, and SimpleNet \cite{liu2023simplenet}
employ an embedding-based framework integrated with anomaly synthesis.
Specifically, DSR and SimpleNet generate anomalies in the feature space,
while the other synthesis-based methods generate anomalies in the image space.

\begin{figure}[t]
  \centering
  \includegraphics[width=0.99\linewidth]{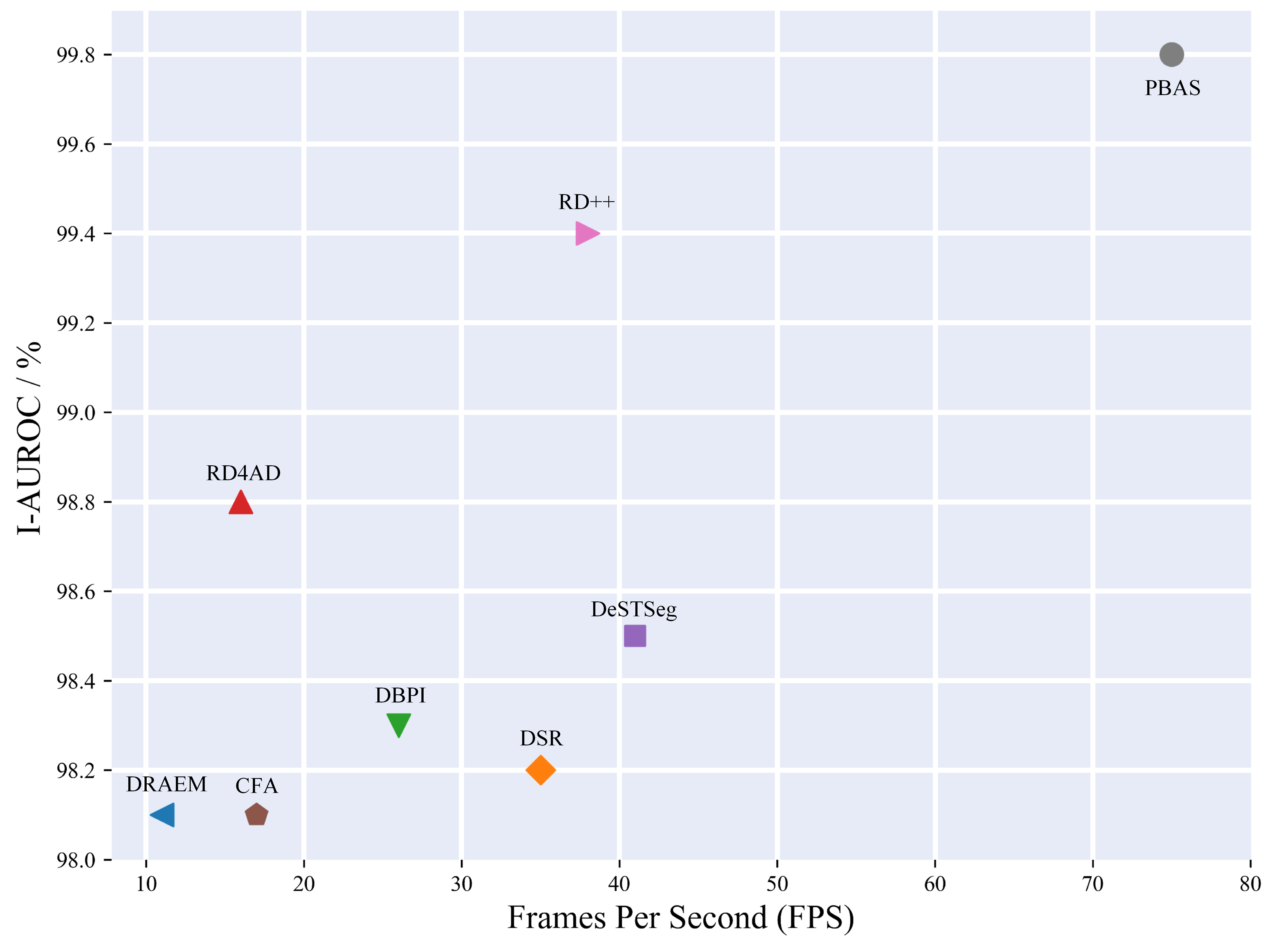}
  \caption{Inference speed (measured in FPS) and detection accuracy (measured by I-AUROC\%) of different methods on MVTec AD.}
  \label{fig:fps}
\end{figure}

\textit{1) Results on MVTec AD:}
As shown in Table~\ref{tab:mvtec_compare_a}, PBAS achieves a perfect I-AUROC of 100\% on 9 categories of MVTec AD,
establishing itself as the SOTA method for anomaly detection with an average I-AUROC of 99.8\%.
It is evident that PBAS excels particularly in the detection of texture categories.
Given that methods using reconstruction-based framework are prone to generating false negatives in difference analysis, 
the average performance of most methods using embedding-based framework surpasses that of methods using reconstruction-based framework.
Through the discriminative learning of synthetic anomalies,
PBAS improves the I-AUROC by a margin of 0.3\% compared to the second-best result
achieved by SimpleNet (which leverages feature-level anomaly synthesis).
As shown in Table~\ref{tab:mvtec_compare_b}, PBAS achieves the best anomaly localization performance, with an average P-AUROC of 98.6\% and P-PRO of 97.3\%.
Specifically, PBAS improves the P-AUROC and P-PRO by margins of 0.3\% and 2.3\% compared to the second-best result, respectively.
Although other methods achieve best performance in several categories, they exhibit significant performance declines in specific categories.

\begin{figure*}[t]
  \centering
  \includegraphics[width=0.99\linewidth]{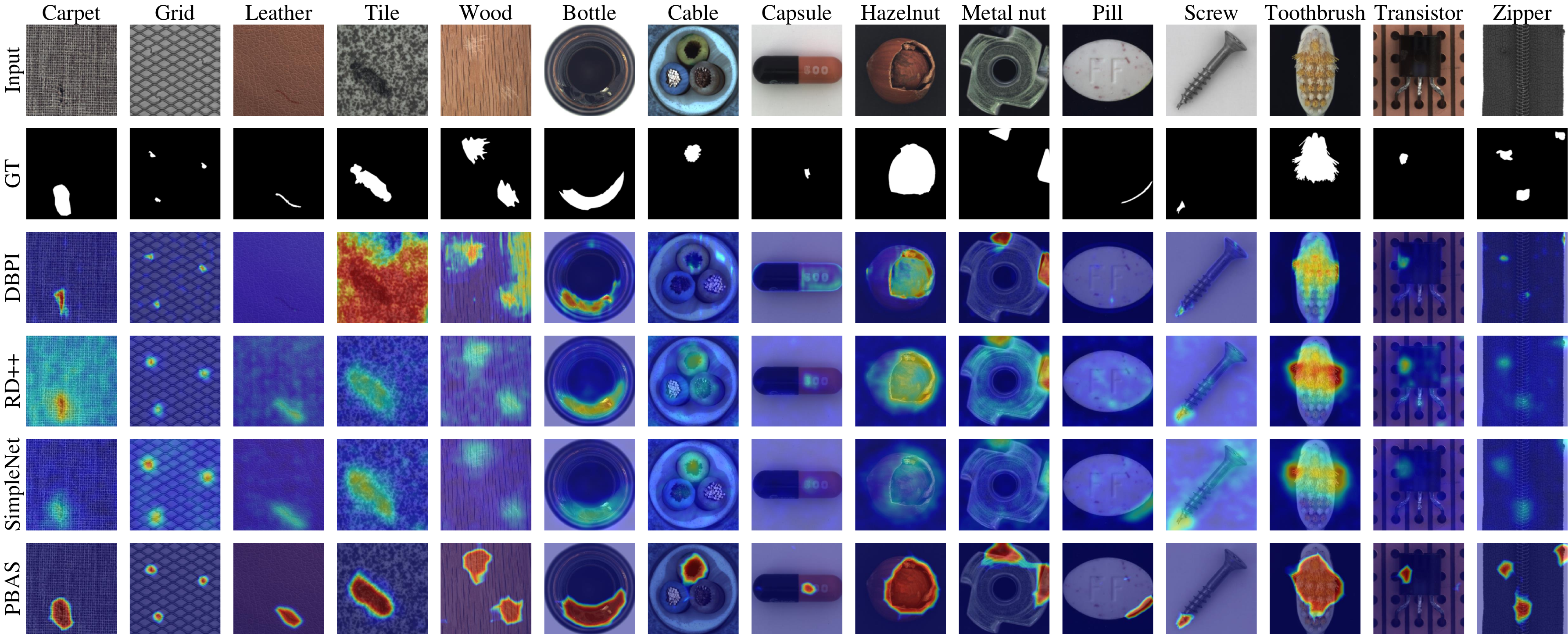}
  \caption{Qualitative comparison of PBAS with various SOTA methods (DBPI \cite{wang2023dual},
  RD++ \cite{tien2023revisiting}, and SimpleNet \cite{liu2023simplenet}) across different categories of the MVTec AD dataset. ``GT'' denotes Ground Truth.}
  \label{fig:mvtec}
\end{figure*}

With a simpler architecture, Fig.~\ref{fig:fps} demonstrates that PBAS achieves a competitive inference speed of 75 FPS,
outperforming these methods in both detection accuracy and efficiency.
As illustrated in Fig.~\ref{fig:mvtec}, PBAS effectively localizes anomalies on each category of MVTec AD.
Due to the uncertain quality of reconstructed image, DBPI \cite{wang2023dual} produces false negatives in many categories.
However, RD++ \cite{tien2023revisiting} and SimpleNet \cite{liu2023simplenet} suffer from low confidence levels.
Compared to these SOTA methods, PBAS completely covers most anomalous regions,
demonstrating its robustness and generalization capability towards unknown anomalies.

\begin{figure*}[t]
  \centering
  \includegraphics[width=0.99\linewidth]{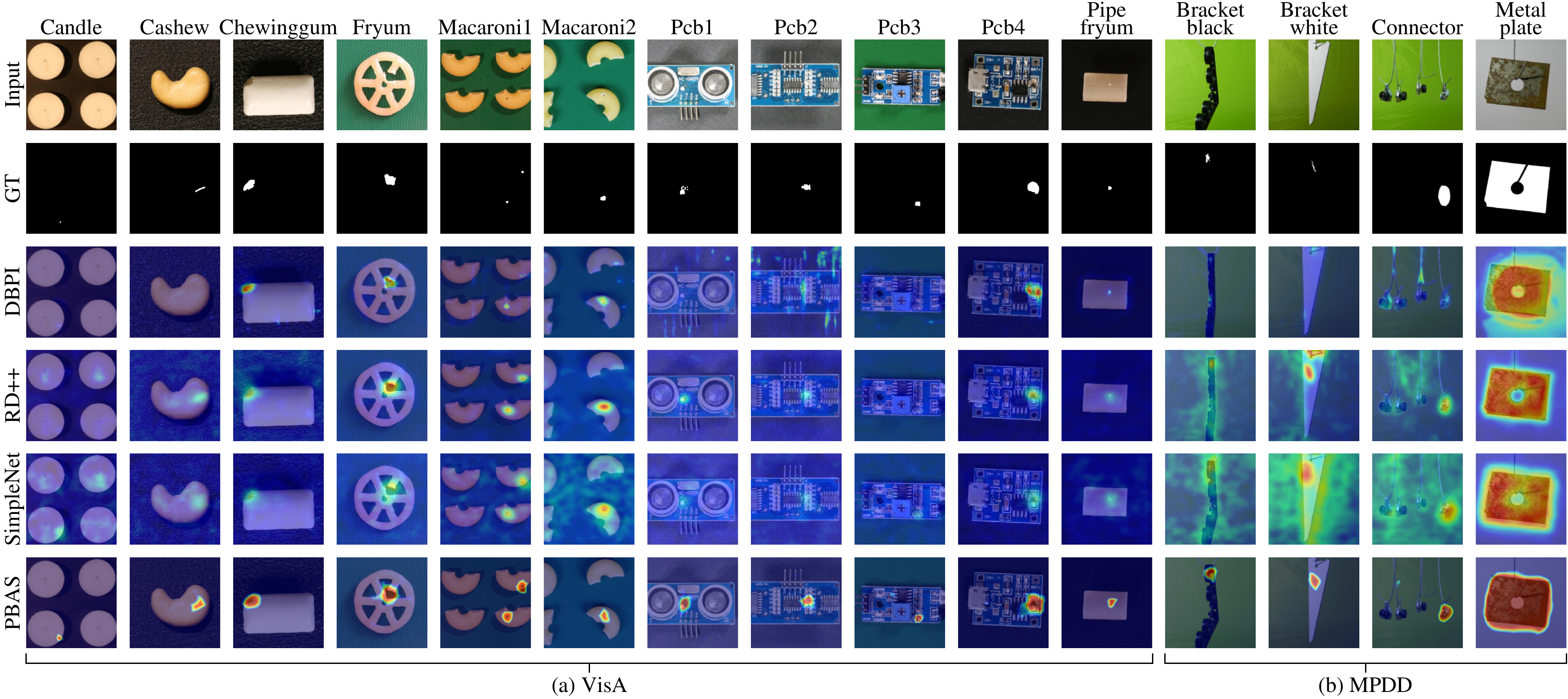}
  \caption{Qualitative comparison of PBAS with various SOTA methods (DBPI \cite{wang2023dual},
  RD++ \cite{tien2023revisiting}, and SimpleNet \cite{liu2023simplenet}) across different categories in the VisA and MPDD datasets. ``GT'' denotes Ground Truth.}
  \label{fig:visa_mpdd}
\end{figure*}

\begin{table*}[t]\scriptsize
  \centering
  \caption{Performance comparison of different methods on VisA, as measured by I-AUROC\%, I-AP\%, P-AUROC\%, P-AP\%, and P-PRO\%.
  The best results for each category are highlighted in bold, and the second-best results are underlined.}
  \begin{tabularx}{0.99\textwidth}{X|*{3}{>{\centering\arraybackslash}X}|*{6}{>{\centering\arraybackslash}X}|>{\centering\arraybackslash}X}
    \hline
    \multirow{3}{*}{Metrics} & \multicolumn{3}{c|}{Embedding-based methods} & \multicolumn{7}{c}{Synthesis-based methods} \\
\cline{2-11}        & NoCoAD & CFA   & RD4AD & DRAEM & DSR   & DBPI  & DeSTSeg & RD++ & SimpleNet  & \multirow{2}{*}{PBAS} \\
          & \cite{gao2023exploring}   & \cite{lee2022cfa}   & \cite{deng2022anomaly}   & \cite{zavrtanik2021draem}  
          & \cite{zavrtanik2022dsr}   & \cite{wang2023dual}   & \cite{zhang2023destseg} & \cite{tien2023revisiting}  & \cite{liu2023simplenet}   &  \\
    \hline
    I-AUROC & 94.4  & 92.0  & 96.0  & 88.7  & 88.0  & 85.4  & 89.8           &  96.3  & \underline{97.1}  & \textbf{97.7} \\
    I-AP  & 94.9  & 93.6  & 96.5  & 90.5  & 91.2  & 87.4  & 90.7             &  96.4 & \underline{97.9}  & \textbf{98.1} \\
    P-AUROC & 98.3  & 84.3  & 90.2  & 93.5  & 84.3  & 90.9  & 97.8           &  \textbf{98.7} & 98.2  & \underline{98.6} \\
    P-AP  & \underline{42.2}  & 26.8  & 27.7  & 26.6  & 35.3  & 33.9  & 40.3 &  42.1 & 37.3  & \textbf{47.6} \\
    P-PRO & 91.8  & 55.1  & 70.9  & 72.4  & 61.9  & 75.3  & 91.3             &   \underline{92.2} & 90.7  & \textbf{93.3} \\
    \hline
  \end{tabularx}%
  \label{tab:visa_compare}%
\end{table*}%

\begin{table*}[t]\scriptsize
  \centering
  \caption{Performance comparison of different methods on MPDD, as measured by I-AUROC\%, I-AP\%, P-AUROC\%, P-AP\%, and P-PRO\%.
  The best results for each category are highlighted in bold, and the second-best results are underlined.}
  \begin{tabularx}{0.99\textwidth}{X|*{3}{>{\centering\arraybackslash}X}|*{6}{>{\centering\arraybackslash}X}|>{\centering\arraybackslash}X}
    \hline
    \multirow{3}{*}{Metrics} & \multicolumn{3}{c|}{Embedding-based methods} & \multicolumn{7}{c}{Synthesis-based methods} \\
\cline{2-11}       & NoCoAD & CFA   & RD4AD & DRAEM & DSR   & DBPI  & DeSTSeg & RD++  & SimpleNet & \multirow{2}{*}{PBAS} \\
          & \cite{gao2023exploring}   & \cite{lee2022cfa}   & \cite{deng2022anomaly}   & \cite{zavrtanik2021draem}  
          & \cite{zavrtanik2022dsr}   & \cite{wang2023dual}   & \cite{zhang2023destseg}  & \cite{tien2023revisiting}  & \cite{liu2023simplenet}   &  \\
    \hline
    I-AUROC & 91.6  & 92.3  & 92.7  & 94.1  & 81.0  & 81.8  & 91.0  & 95.5                         & \textbf{98.1} & \underline{97.7} \\
    I-AP  & 93.8  & 92.2  & 95.3  & 96.1  & 87.2  & 85.4  & 93.0  & 96.7                           & \textbf{98.7} & \underline{98.0} \\
    P-AUROC & 98.3  & 94.8  & \underline{98.7}  & 91.8  & 76.2  & 92.3  & 94.1  & \underline{98.7} & \underline{98.7} & \textbf{98.8} \\
    P-AP  & 36.8  & 28.3  & \textbf{45.6} & 28.8  & 21.5  & 21.5  & 23.6  & 34.4                   & 32.5 & \underline{37.8} \\
    P-PRO & 94.9  & 83.2  & 95.4  & 78.2  & 58.4  & 85.6  & 83.3  & 95.6                           & \underline{95.7} & \textbf{97.1} \\
    \hline
  \end{tabularx}%
  \label{tab:mpdd_compare}%
\end{table*}%

\textit{2) Results on VisA:} 
As shown in Table~\ref{tab:visa_compare}, PBAS achieves superior anomaly detection and localization performance on VisA,
with an average I-AUROC of 97.7\%, P-AUROC of 98.6\%, and P-PRO of 93.3\%.
Specifically, PBAS improves the I-AUROC, I-AP, P-AP, and P-PRO by margins of 0.6\%, 0.2\%, 5.4\%, and 1.1\% compared to the second-best result, respectively.
As illustrated in Fig.~\ref{fig:visa_mpdd}(a), PBAS effectively localizes anomalies on different categories within VisA.
Due to the relatively small size of some anomalies in VisA (e.g., the ``Candle'' class), DBPI, RD++, and SimpleNet frequently exhibit incorrect localization.
The results demonstrate that PBAS excels in detecting tiny anomalies.
Given that P-AP is particularly sensitive to small-scale anomalies, PBAS outperforms other methods significantly on this metric.
Similarly, the average I-AUROC and I-AP of embedding-based framework surpass those of reconstruction-based framework.

\textit{3) Results on MPDD:} 
As shown in Table~\ref{tab:mpdd_compare}, PBAS achieves competitive anomaly detection and localization performance on MPDD,
with an average I-AUROC of 97.7\%, P-AUROC of 98.8\%, and P-PRO of 97.1\%.
Specifically, PBAS improves the P-AUROC and P-PRO by margins of 0.1\% and 1.4\% compared to the second-best result, respectively.
As illustrated in Fig.~\ref{fig:visa_mpdd}(b), PBAS effectively localizes anomalies on different categories within MPDD.
Due to the complex camera conditions in MPDD (e.g., the ``Bracket black'' class), RD++ and SimpleNet are prone to over-detection,
while DBPI fails to detect anomalies.
The results demonstrate that PBAS excels in detecting anomalies across various shooting angles.
Given that P-AUROC is insensitive to over-detection, PBAS shows only a slight improvement over others.
Similarly, the average P-AUROC and P-PRO of embedding-based framework surpass those of reconstruction-based framework.

\subsection{Ablation Study}
\label{sec:_ablation}

To verify the effectiveness of different components in PBAS, we conduct ablation experiments on MVTec AD.
PBAS consists of three core components:
Approximate Boundary Learning (ABL), Anomaly Feature Synthesis (AFS), and Refined Boundary Optimization (RBO).

\begin{table}[t]\small
  \centering
  \caption{Performance ablation of different components in PBAS on MVTec AD, as measured by I-AUROC\%, P-AUROC\%, and P-PRO\%.
  ``${\cal A}$'', ``${\cal F}$'', ``${\cal N}$'', and ``${\cal S}$'' stand for
  Average center method, Feature alignment method, Noisy anomaly synthesis, and Self-adaptive generation, respectively.
  The best results are highlighted in bold.
  }
    \begin{tabularx}{0.475\textwidth}{*{4}{>{\centering\arraybackslash}p{0.25cm}|}>{\centering\arraybackslash}p{0.6cm}|*{3}{>{\centering\arraybackslash}X}}
    \hline
    \multicolumn{2}{c|}{ABL} & \multicolumn{2}{c|}{AFS} & \multirow{2}{*}{RBO} & \multirow{2}{*}{I-AUROC} & \multirow{2}{*}{P-AUROC} & \multirow{2}{*}{P-PRO} \\
    \cline{1-4}      ${\cal A}$ & ${\cal F}$  & ${\cal N}$ & ${\cal S}$  &       &       &       &  \\
    \hline
    \ding{51}     &       &       &       &       & 96.3  & 97.6  & 93.2 \\
          & \ding{51}     &       &       &       & 97.9  & 98.1  & 94.2 \\
          & \ding{51}     & \ding{51}     &       & \ding{51}     & 99.0  & 98.1  & 91.9 \\
          & \ding{51}     &       & \ding{51}     & \ding{51}     & \textbf{99.8} & \textbf{98.6} & \textbf{97.3} \\
    \hline
    \end{tabularx}%
  \label{tab:component_ablation}%
\end{table}%

\textit{1) Components in PBAS:}
As shown in Table~\ref{tab:component_ablation}, PBAS achieves the best performance on three metrics when all three components are integrated.
Specifically, we first divide the center initialization method in ABL into two types:
the traditional average center method ${\cal A}$ and our proposed feature alignment method ${\cal F}$.
Since there is no discriminator when using ABL alone,
we define the anomaly score by the Euclidean distance between the anomaly feature and its nearest center.
Due to the intra-class variations, the performance of ABL using method ${\cal F}$ is significantly better than using method ${\cal A}$,
which will be detailed in the next section.
Compared to ABL using method ${\cal A}$, ABL using method ${\cal F}$ improves the I-AUROC, P-AUROC, and P-PRO by margins of 1.6\%, 0.5\%, and 1.0\%, respectively.

Next, we divide the anomaly synthesis strategy in AFS into two types:
noisy anomaly synthesis ${\cal N}$ and our proposed self-adaptive generation method ${\cal S}$.
Similar to \cite{liu2023simplenet}, method ${\cal N}$ synthesizes anomalies by adding Gaussian noise to normal features.
Based on ABL using method ${\cal F}$ and integrated with the discriminative learning of RBO,
the performance of AFS using method ${\cal S}$ is significantly better than using method ${\cal N}$. 
Compared to AFS using method ${\cal N}$, AFS using method ${\cal S}$ improves
the I-AUROC, P-AUROC, and P-PRO by margins of 0.8\%, 0.5\%, and 5.4\%, respectively.
Due to the fixed variance and random directions of Gaussian noise,
the anomalies generated by AFS using method ${\cal N}$ are scale-invariant and lack directionality.
On the other hand, the anomalies generated by AFS using method ${\cal S}$ are located on the ray direction from the center to normal features.
As the anomalies progressively contract with the increase of training epochs, they become more controllable and meaningful.
Finally, through the discriminative learning of normal and anomaly feature in RBO,
the decision boundary transitions from approximate to refined.
Compared to using ABL alone, the introduction of AFS and RBO can further improve the anomaly detection performance.

\begin{figure*}[t]
  \centering
  \includegraphics[width=0.99\linewidth]{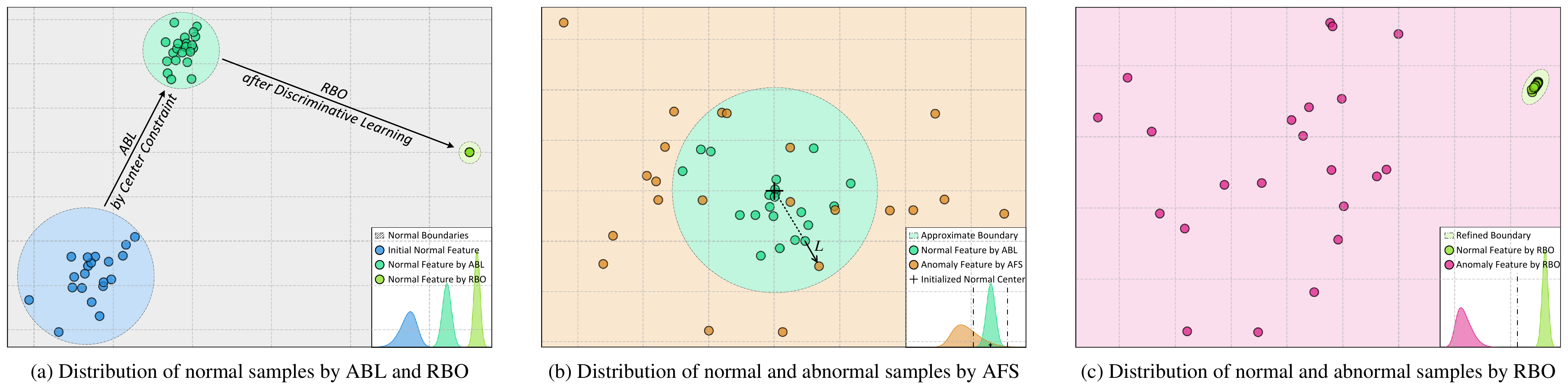}
  \caption{Feature distribution and probability density at different stages reduced by PCA on the ``Leather'' class within MVTec AD.}
  \label{fig:distribution}
\end{figure*}

\textit{2) Visualized Feature Distribution:}
To more intuitively demonstrate the effect of different components on the decision boundary,
we present the actual normal and anomaly feature distributions reduced by Principal Component Analysis (PCA).
As shown in Fig.~\ref{fig:distribution}(a), the normal feature distribution becomes increasingly concentrated
following the center constraint of ABL and the discriminative learning of RBO,
resulting in progressively more compact boundaries.
As shown in Fig.~\ref{fig:distribution}(b), the anomaly features generated by AFS are mixed with the normal features in the hypersphere.
As shown in Fig.~\ref{fig:distribution}(c), the normal and anomaly features learned by RBO are further distinguished.

\begin{figure*}[t]
  \centering
  \includegraphics[width=0.99\linewidth]{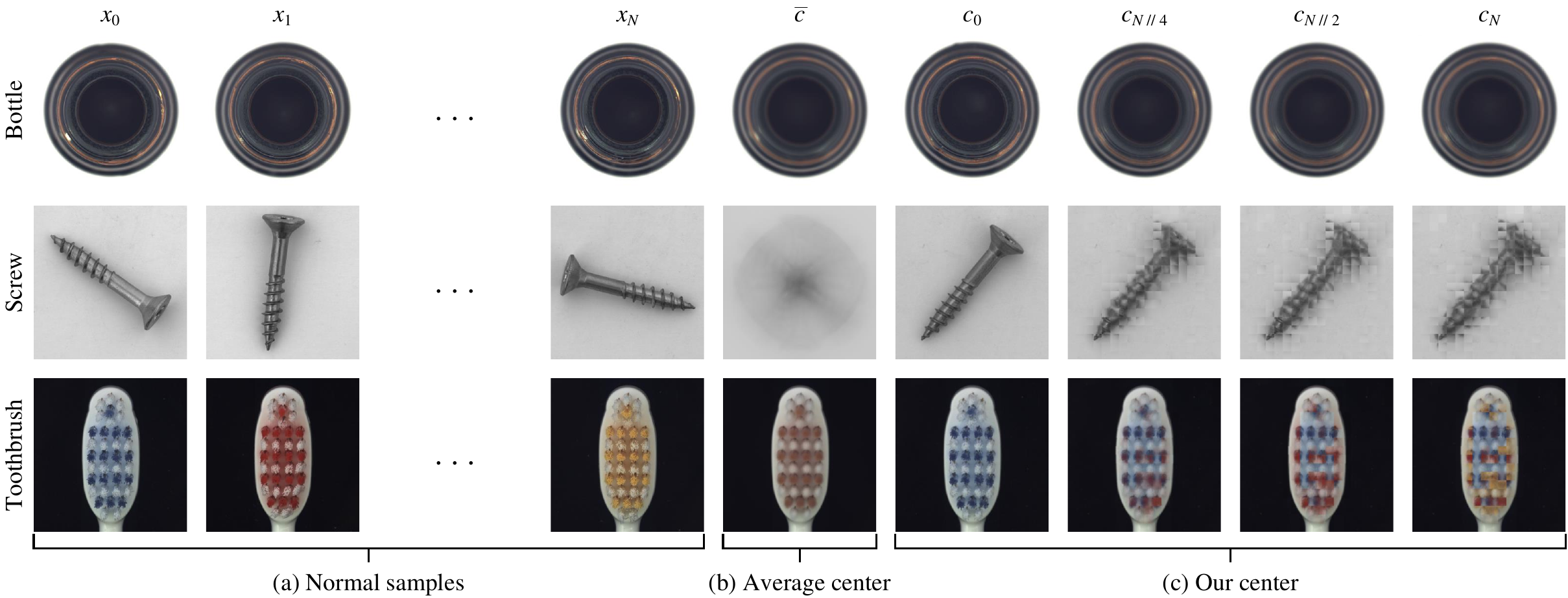}
  \caption{Visualization of center initialization methods on several categories within MVTec AD. (a) Several normal samples from MVTec AD.
  (b) Corresponding average centers by traditional method. (c) Process of our center initialization method.
  $N$ is the total number of normal samples in the training set.}
  \label{fig:center_visual}
\end{figure*}

\textit{3) Components in Center Initialization:}
To further verify the effectiveness of our center initialization method,
we conduct detailed experiments on several such methods using both ABL alone and the entire PBAS.
As shown in Table~\ref{tab:center_ablation}, we have selected four center initialization methods for comparison,
including traditional average center, EMA iterative center, k-means clustering center, and our proposed method.
When using ABL alone, our method improves the I-AUROC, P-AUROC, and P-PRO by margins of 0.8\%, 0.1\%, and 0.7\% compared to the second-best result, respectively.
Through the feature alignment with iterative updates,
our method more accurately determines the feature center, thereby yielding significant improvements.
When using the entire PBAS, all four methods achieve higher performance.
However, our method further improves the P-PRO by a margin of 0.6\%.

\begin{table}[t]\small
  \centering
  \caption{Performance ablation of different center initialization methods in ABL and PBAS on MVTec AD, as measured by I-AUROC\%, P-AUROC\%, and P-PRO\%.
  The best results are highlighted in bold.}
    \begin{tabularx}{0.475\textwidth}{>{\centering\arraybackslash}p{0.75cm}|>{\raggedright\arraybackslash}p{1.2cm}|*{3}{>{\centering\arraybackslash}X}}
    \hline
    \multicolumn{2}{c|}{Methods} & I-AUROC & P-AUROC & P-PRO \\
    \hline
    \multirow{4}{*}{ABL} & Average & 96.3  & 97.6  & 93.2 \\
    \cline{2-2}          & Iterative & 97.1  & 98.0  & 93.5 \\
    \cline{2-2}          & Clustering & 96.4  & 97.6  & 93.3 \\
    \cline{2-2}          & Ours  & \textbf{97.9} & \textbf{98.1} & \textbf{94.2} \\
    \hline
    \multirow{4}{*}{PBAS} & Average & 99.7  & 98.5  & 96.7 \\
    \cline{2-2}          & Iterative & 99.7  & 98.5  & 96.4 \\
    \cline{2-2}          & Clustering & 99.5  & 98.3  & 96.0 \\
    \cline{2-2}          & Ours  & \textbf{99.8} & \textbf{98.6} & \textbf{97.3} \\
    \hline
    \end{tabularx}%
  \label{tab:center_ablation}%
\end{table}%

To further evaluate the impact of the frozen feature projector $P_\theta$ during center initialization,
we conduct an ablation study on the MVTec AD dataset.
The results indicate that the model with the feature projector significantly outperforms the model without it.
Specifically, PBAS without $P_\theta$ achieves an average I-AUROC of 99.6\%, which is a decrease of 0.2\% compared to PBAS with $P_\theta$.
These findings demonstrate that $P_\theta$ produces a Gaussian filter-like effect on features,
stabilizing center initialization and enhancing the model's ability to effectively capture normal patterns.

\textit{4) Iterative Process of Center Initialization:}
To more intuitively demonstrate the superiority of our center initialization method,
Fig.~\ref{fig:center_visual} provides the visualization of different centers at the image level.
The average center in Fig.~\ref{fig:center_visual}(b) is given by the normal samples in Fig.~\ref{fig:center_visual}(a) through Eq.~\ref{eq:traditional}.
Since there is little variation among samples in the ``Bottle'' class,
the average center can effectively represent the normal samples.
However, due to positional variations in the ``Screw'' class, the average center fails to represent the normal samples.
Although there are no positional variations in the ``Toothbrush'' class, there are noticeable color variations.
Consequently, the average center method cannot correctly handle situations with intra-class variations.
Fig.~\ref{fig:center_visual}(c) illustrates the iterative update process of our method based on feature alignment.
When intra-class variations are negligible (such as in the ``Bottle'' class),
our method accurately captures key information and provides clearer results.
Meanwhile, when intra-class variations are obvious (such as in the ``Screw'' and ``Toothbrush'' classes),
our method can still cover all key information.

\begin{figure}[t]
  \centering
  \includegraphics[width=0.99\linewidth]{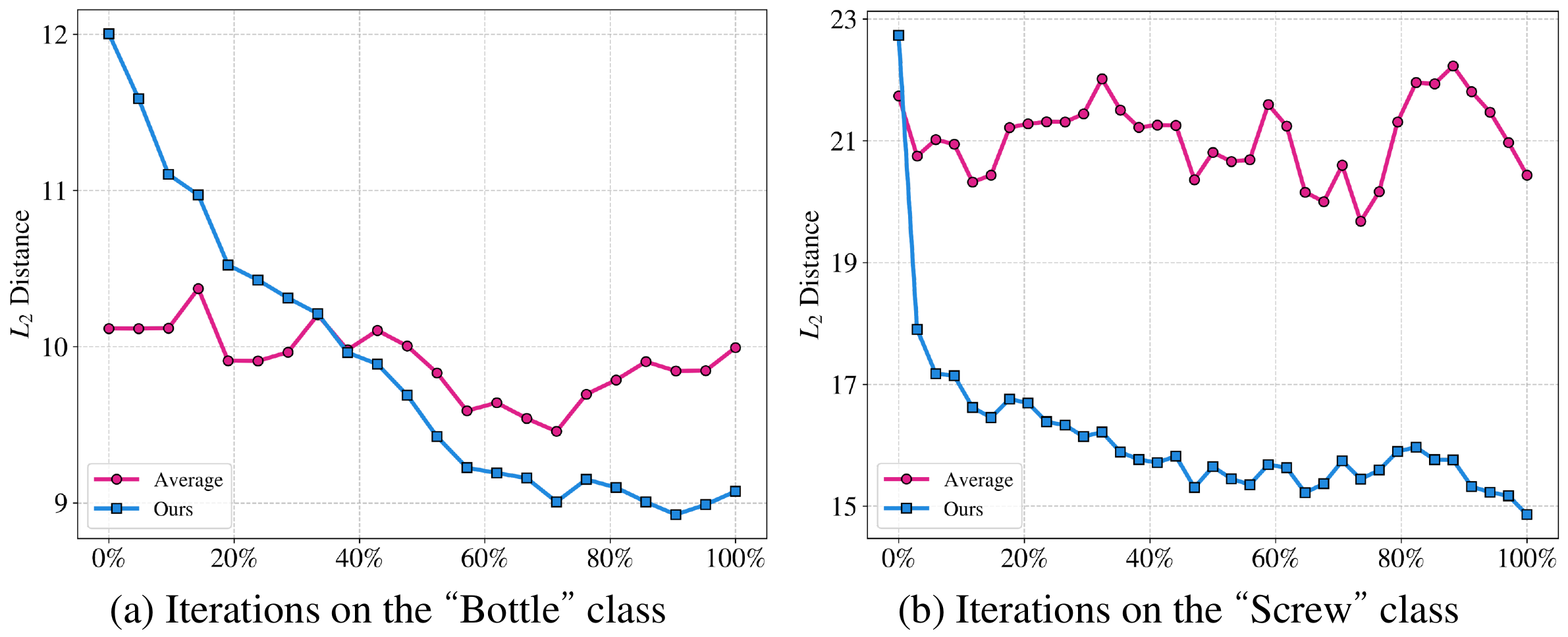}
  \caption{Quantitative comparison of the \(L_2\) distance between samples and different centers
  for two classes in the MVTec AD dataset during batch iterations of center initialization.}
  \label{fig:distance}
\end{figure}

To assess the effectiveness of our center initialization,
we analyze the \(L_2\) distance between samples and their corresponding centers during batch iterations.
A smaller distance means better capture of normal patterns.
As shown in Fig.~\ref{fig:distance},
our proposed center progressively captures normal patterns over the iterations, while the average center remains mostly unchanged.
For the ``Bottle'' class with less intra-class variation in Fig.~\ref{fig:distance}(a),
our center shows a slight improvement over the average center.
However, for the ``Screw'' class with higher intra-class variation in Fig.~\ref{fig:distance}(b),
our center significantly outperforms the average center.
As a result, when using ABL alone on the ``Screw'' class,
our method achieves a 17.1\% increase (from 71.3\% to 88.4\%) in I-AUROC compared to the average center method.
Given the convergence of our method across categories, the experiments utilize all samples for center initialization.
In practical scenarios, early stopping based on the percentage decrease in \(L_2\) distance can be applied to ensure efficient convergence.
In summary, our center initialization method can more effectively represent normal samples, resulting in superior performance.

\subsection{Parameter Analyses}
\label{sec:_parameter}

To further explore how different hyperparameters affect the performance of PBAS,
we carry out five independent parameter analyses on MVTec AD.
During the analysis of a certain parameter, all other parameters are using the default settings,
which yield the best results in corresponding analyses.

\begin{figure}[t]
  \centering
  \includegraphics[width=0.99\linewidth]{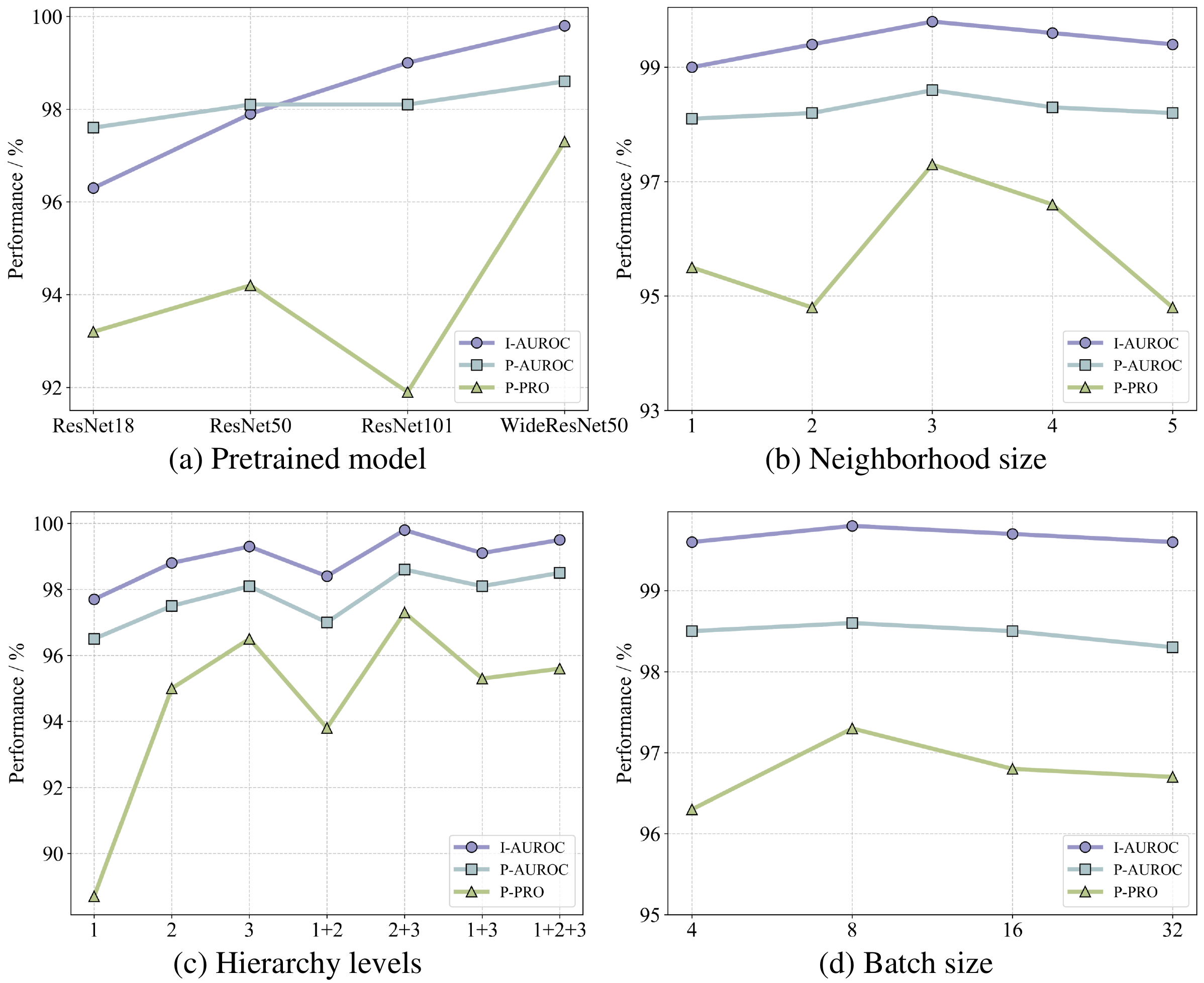}
  \caption{Quantitative results of several parameter analyses on MVTec AD, as measured by I-AUROC\%, P-AUROC\%, and P-PRO\%. 
  }
  \label{fig:parameter}
\end{figure}

\textit{1) Dependence on pretrained Model:}
In our proposed method PBAS, the first component ABL uses a pretrained model to extract raw features.
To explore the dependency of PBAS on different pretrained networks,
we compare four ResNet-like backbones with increasing model complexity:
ResNet18, ResNet50, ResNet101 \cite{he2016deep}, and WideResNet50 \cite{zagoruyko2016wide}.
As shown in Fig.~\ref{fig:parameter}(a), the overall anomaly detection and localization performance is positively correlated with the number of parameters.
Specifically, WideResNet50 improves the I-AUROC, P-AUROC, and P-PRO by margins of 0.8\%, 0.5\%, and 3.1\% compared to the second-best result, respectively.
Therefore, WideResNet50 is chosen as the default setting for backbone $\phi$ of feature extractor ${{E}_{\phi}}$ in ABL.
In this paper, all comparative methods using the embedding-based framework adopt WideResNet50 as the backbone.

\textit{2) Selection of Neighborhood Size:}
To enhance robustness against small spatial deviations,
the neighborhood size $p$ is utilized as the patch size for feature aggregation in Eq.~\ref{eq:neighbour_location}.
Fig.~\ref{fig:parameter}(b) indicates that the best result is achieved when \mbox{$p=3$}.
Specifically, \mbox{$p=3$} improves the I-AUROC, P-AUROC, and P-PRO by margins of 0.2\%, 0.3\%, and 0.7\% compared to the second-best result, respectively.
If $p$ is too small, the feature aggregation is insufficient, leading to a lack of spatial information.
Conversely, if $p$ is too large, the feature aggregation is excessive, resulting in a loss of detailed information.

\textit{3) Concatenation of Hierarchy Levels:} 
As introduced in Section~\ref{sec:_approximate}, the hierarchy levels of aggregated features are concatenated to
capture low-level and high-level features in Eq.~\ref{eq:final_feature}.
Fig.~\ref{fig:parameter}(c) indicates that the best result is obtained by concatenating the features from hierarchy levels 2 and 3.
Specifically, the concatenation of the two levels improves the I-AUROC, P-AUROC, and P-PRO 
by margins of 0.3\%, 0.1\%, and 0.8\% compared to the second-best result, respectively.
It is evident that the three metrics follow a similar trend.
Since hierarchy level 3 contains the deepest semantic information, this level contributes the most to the improvement of performance.

\textit{4) Influence of Batch Size:}
In addition to impacting the convergence speed,
the batch size also affects our center initialization method in ABL.
Fig.~\ref{fig:center} demonstrates that the batch size influences the number of samples averaged in each iteration.
Meanwhile, the batch size serves as a latent factor that influences the range of anomaly synthesis in AFS.
Specifically, the anomaly synthesis process in Eq.~\ref{eq:synthesis} is constrained by the flexible length \({\cal L}_{\text{c}}\),
which is the loss function of ABL influenced by the batch size in Eq.~\ref{eq:l_center}.
As shown in Fig.~\ref{fig:parameter}(d), the variation of three metrics is not significant,
indicating that PBAS is robust with respect to the batch size.
Experimentally, the best result is achieved when the batch size is set to 8.

\begin{figure}[t]
  \centering
  \includegraphics[width=0.99\linewidth]{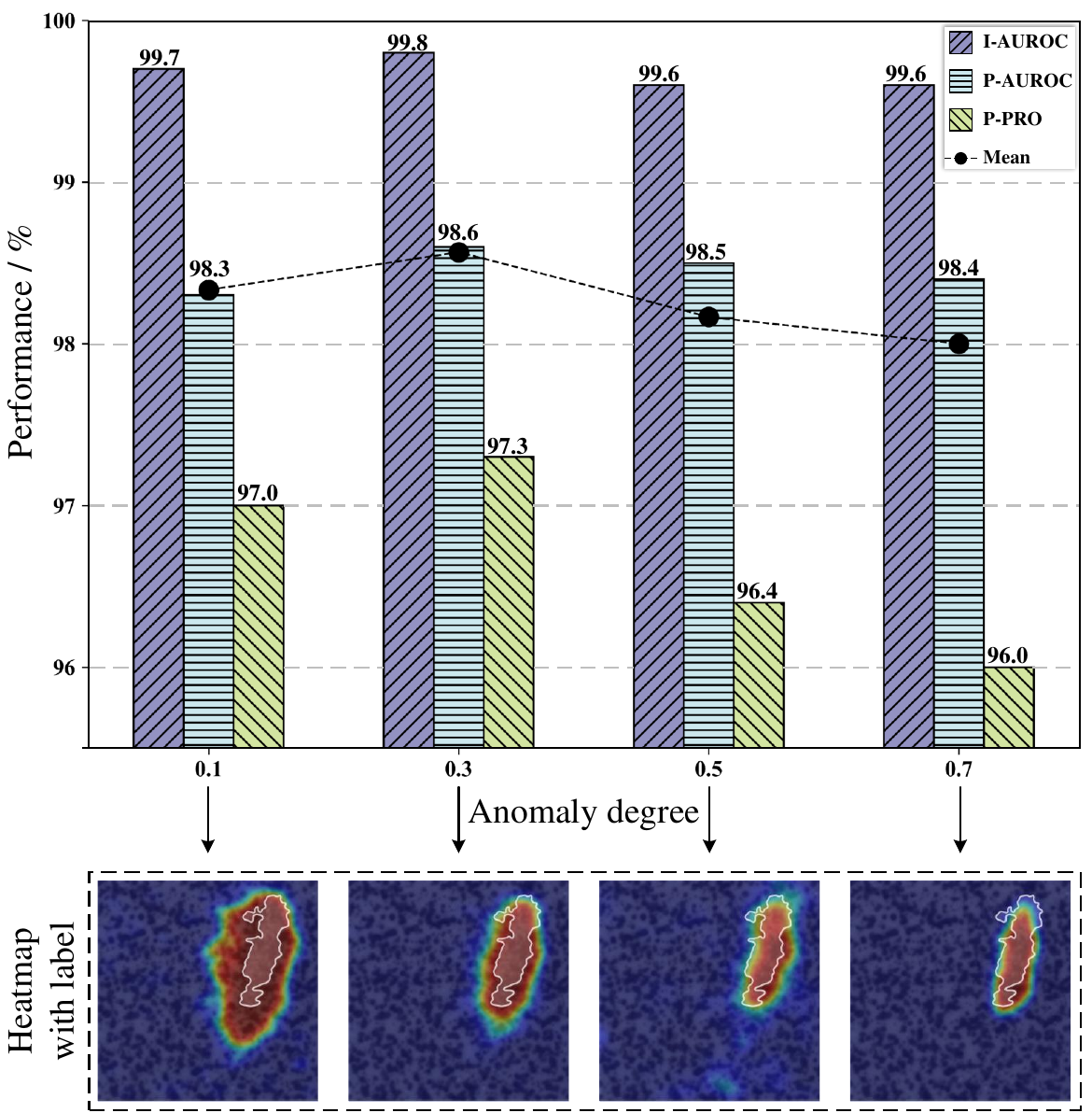}
  \caption{Quantitative and qualitative results of different anomaly degrees $\alpha$ on MVTec AD, as measured by I-AUROC\%, P-AUROC\%, and P-PRO\%.
  Corresponding heatmaps with white labels are shown above, indicating the visualization of anomaly localization.}
  \label{fig:synthesis}
\end{figure}

\textit{5) Impact of Anomaly Degree:}
We have discussed the batch size as a latent factor that influences the range of anomaly synthesis.
In contrast, the anomaly degree $\alpha$ is an explicit factor.
As mentioned in Section~\ref{sec:_synth}, $\alpha$ is proposed to control the range of anomaly synthesis in AFS.
As $\alpha$ decreases, the synthetic anomalies become closer to the normal features, and vice versa.
The upper part of Fig.~\ref{fig:synthesis} indicates that the best result is achieved when \mbox{\(\alpha=0.3\)}.
Specifically, \mbox{\(\alpha=0.3\)} improves the I-AUROC, P-AUROC, and P-PRO by margins of 0.1\%, 0.1\%, and 0.3\% compared to the second-best result, respectively.
As shown in the lower part of Fig.~\ref{fig:synthesis}, the area of the detected anomaly regions gradually decrease with the increase of $\alpha$.
This implies that a lower $\alpha$ is more effective in detecting subtle anomalies,
while a higher $\alpha$ emphasizes the purity of detection in normal regions.
To better balance between the over-detection and under-detection,
we choose \mbox{\(\alpha=0.3\)} as the default setting for anomaly synthesis.

\begin{table*}[htbp]\small
    \centering
    \caption{Quantitative comparison of SimpleNet and PBAS with different subsampling ratios on MVTec AD,
    as measured by I-AUROC\% and P-AUROC\%. The best results are highlighted in bold.}
    \begin{tabularx}{\textwidth}{>{\hsize=55pt}X|
        *{5}{>{\centering\arraybackslash}X}|
        *{5}{>{\centering\arraybackslash}X}}
      \hline
      \multirow{2}{*}{Metrics} & \multicolumn{5}{c|}{SimpleNet \cite{liu2023simplenet}}   & \multicolumn{5}{c}{PBAS} \\
  \cline{2-11}       & 20\%  & 40\%  & 60\%  & 80\%  & 100\% & 20\%  & 40\%  & 60\%  & 80\%  & 100\% \\
      \hline
      I-AUROC & 95.7  & 98.1  & 98.5  & 99.3  & \textbf{99.5} & 96.9  & 98.4  & 98.9  & 99.7  & \textbf{99.8} \\
      P-AUROC & 97.4  & 97.7  & 97.8  & 97.9  & \textbf{98.1} & 98.1  & 98.3  & 98.4  & 98.5  & \textbf{98.6} \\
      \hline
    \end{tabularx}%
    \label{tab:subsampling}%
\end{table*}%

\textit{6) Subsampling of Training Data:}
In cases with a limited number of training samples,
the center feature may capture fewer normal patterns.
As shown in Table~\ref{tab:subsampling},
although the performance decreases as the subsampling ratio is reduced,
PBAS maintains high I-AUROC and P-AUROC performance even at the lowest subsampling ratio of 20\%.
Specifically, PBAS shows a 2.9\% decline in I-AUROC, compared to a 3.8\% decline for SimpleNet \cite{liu2023simplenet}.
Therefore, PBAS not only consistently outperforms SimpleNet
across all subsampling ratios but also exhibits greater robustness to sample size reduction.
This robustness is attributed to our center initialization, which employs iterative updates and feature alignment.

\begin{table}[t]\small
  \centering
  \caption{Quantitative results of different initial center $c_0$ on MVTec AD, as measured by I-AUROC\%, P-AUROC\%, and P-PRO\%.}
  \begin{tabularx}{0.4\textwidth}{>{\hsize=20pt}X|*{3}{>{\centering\arraybackslash}X}}
    \hline
    Seed  & I-AUROC & P-AUROC & P-PRO \\
    \hline
    0     & 99.8 & 98.6 & 97.3 \\
    1     & 99.7  & 98.6 & 97.2 \\
    2     & 99.8 & 98.5  & 97.4 \\
    3     & 99.7  & 98.6 & 97.3 \\
    4     & 99.8 & 98.5  & 97.2 \\
    \hline
    \end{tabularx}%
  \label{tab:center_select}%
\end{table}%

\textit{7) Sensitivity to Initial Center Variations:}
Since our center initialization method iteratively updates from the first batch,
the randomness in the selection of samples in initial center $c_0$ needs to be considered.
As shown in Table~\ref{tab:center_select}, the performance of PBAS remains relatively stable across different seeds.
This is because we divide the dataset into several batches and process them iteratively,
allowing the center feature to capture intra-class diversity and reducing reliance on the initial batch.
By employing an EMA for updating, the center retains past information while incorporating new data,
enabling the model to converge to a more robust representation of the normal feature space.
As a result, PBAS is not sensitive to variations in the initial center.
\section{Conclusion}
\label{sec:conclu}

In this paper, we propose a novel feature-level anomaly synthesis strategy guided
by the progressive boundary for enhancing anomaly detection, termed PBAS.
Our method addresses key limitations of existing approaches by eliminating the need for
predefined anomaly properties and allowing for controllable anomaly synthesis. 
Leveraging the hypersphere boundary established by ABL and the artificial anomalies synthesized by AFS,
RBO refines the boundary between normal and anomalous samples through binary classification.
Consequently, PBAS significantly improves the performance of anomaly detection and localization.
We evaluate our method on various industrial datasets and achieve state-of-the-art performance,
demonstrating the effectiveness and efficiency of PBAS.
Moreover, PBAS has the potential to detect subtle anomalies.
Since PBAS is designed to capture and enhance distribution differences in the feature space through anomaly synthesis and boundary optimization,
our main focus is localizing structural anomalies (e.g., surface stains) in industrial scenarios.
However, logical anomalies (e.g., misassembled parts),
which often require higher-level understanding such as scene interpretation,
are beyond the current scope of PBAS.
We have not yet thoroughly addressed these types of anomalies.
In the future, we will explore the integration of semantic analysis for logical anomaly detection.

\bibliography{main}
\bibliographystyle{IEEEtran}

\begin{IEEEbiography}[{\includegraphics[width=0.99in,height=1.25in,clip,trim=0cm 0.2cm 0cm 0.2cm]{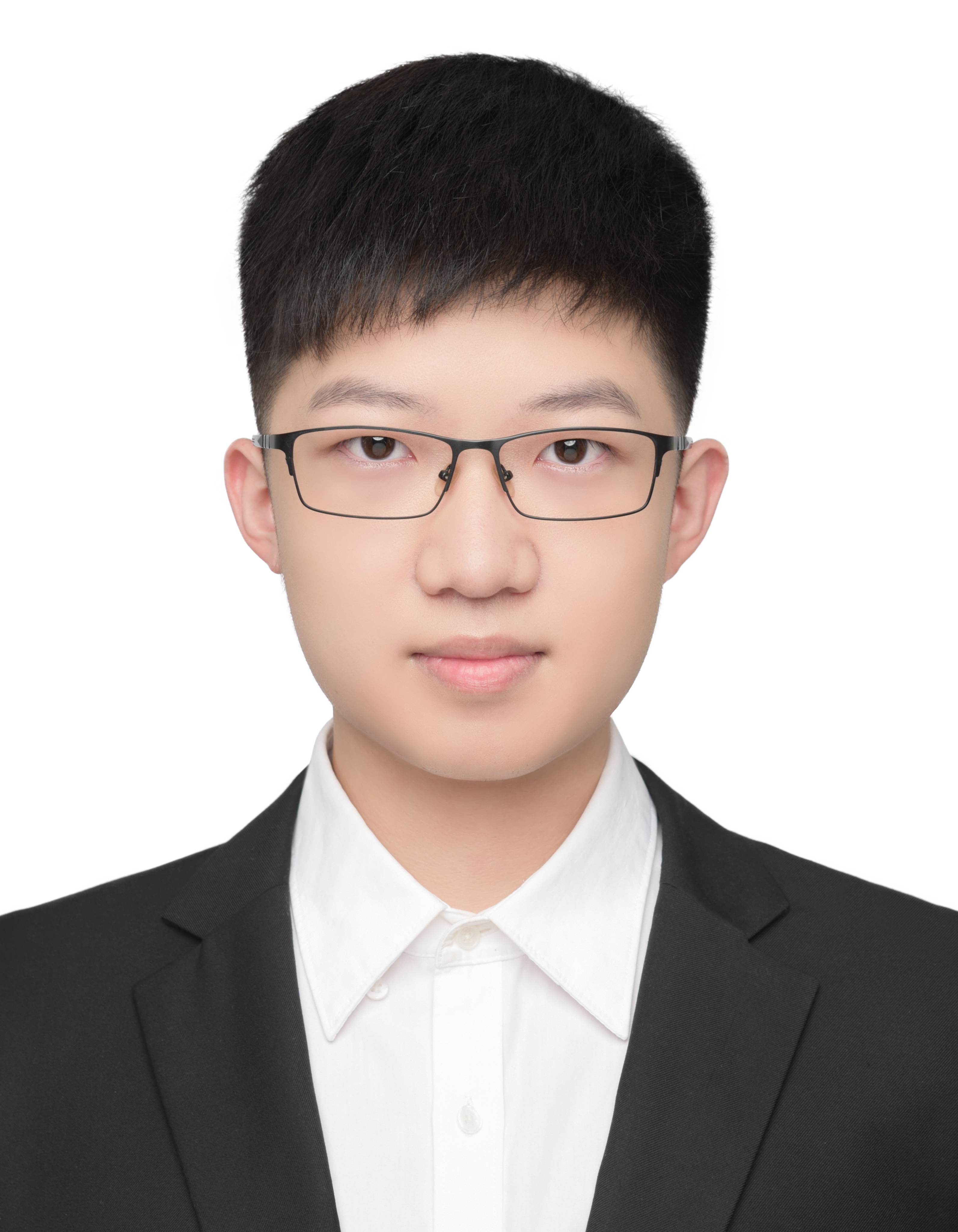}}]{Qiyu Chen}
    received the B.E. degree in Communication Engineering from Tongji University, Shanghai, China, in 2021.
    He is currently pursuing the Ph.D. degree with the Institute of Automation, Chinese Academy of Sciences (IACAS), Beijing, China,
    and also with the School of Artificial Intelligence, University of Chinese Academy of Sciences, Beijing.

    His research interests include deep learning, computer vision, and anomaly detection.
\end{IEEEbiography}

\begin{IEEEbiography}[{\includegraphics[width=0.99in,height=1.25in,clip,trim=0cm 0.5cm 0cm 0.5cm]{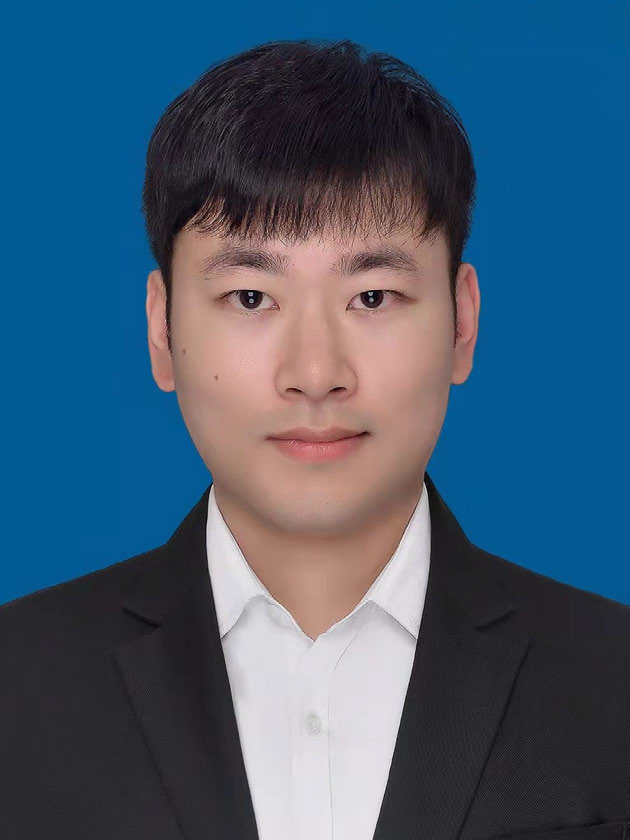}}]{Huiyuan Luo}
    received the B.S. degree from Harbin Institute of Technology, Weihai, China, in 2016, 
    and the Ph.D. degree from the Changchun Institute of Optics, Fine Mechanics and Physics, Chinese Academy of Science, Changchun, China, in 2021.
    Since 2022, he has been a postdoctor and assistant researcher at the Institute of Automation, Chinese Academy of Sciences (IACAS), Beijing.

    He has been engaged in saliency detection, industrial anomaly detection, unsupervised learning, and intelligent manufacturing.
\end{IEEEbiography}

\begin{IEEEbiography}[{\includegraphics[width=0.99in,height=1.25in,clip,trim=10cm 0cm 10cm 0cm]{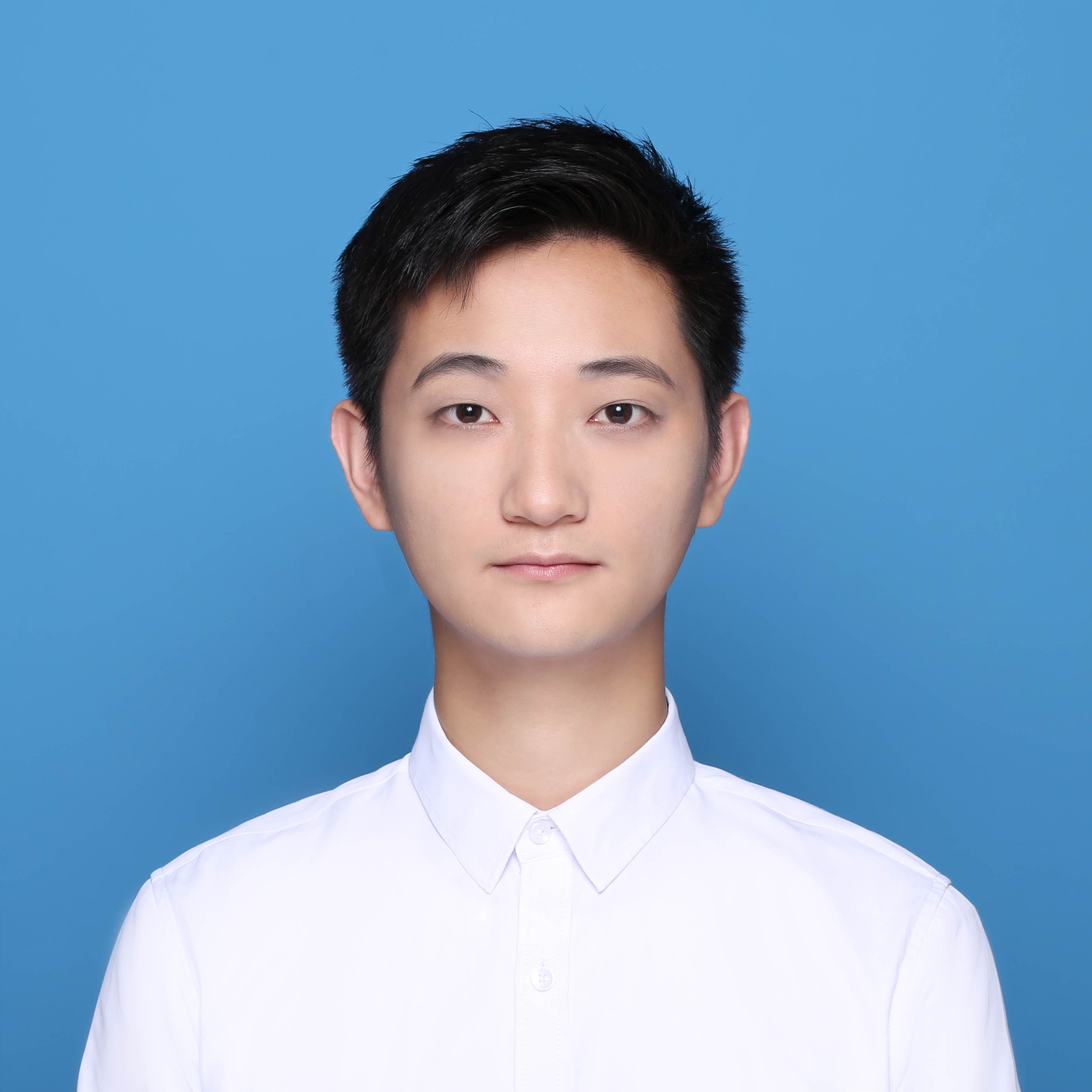}}]{Han Gao}
    received the B.S. degree from Jilin University, Changchun, China, in 2019, and the M.S. degree in control science and engineering from
    the Institute of Automation, Chinese Academy of Sciences (IACAS), Beijing, in 2022.
    He is currently a Ph.D. candidate at the Research Center of Precision Sensing and Control, IACAS,
    and also with the School of Artificial Intelligence, University of Chinese Academy of Sciences, Beijing.

    His research interests include machine learning and visual inspection.
\end{IEEEbiography}

\begin{IEEEbiography}[{\includegraphics[width=0.99in,height=1.25in,clip,trim=0cm 0cm 0cm 0cm]{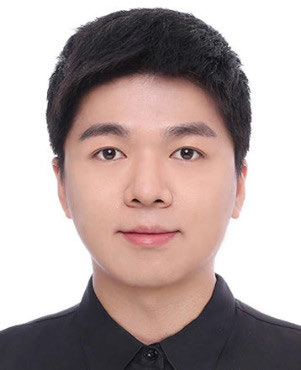}}]{Chengkan Lv}
    (Member, IEEE)
    received the B.S. degree from Shandong University, Jinan, China, in 2017, and the Ph.D. degree
    from the Institute of Automation, Chinese Academy of Sciences, Beijing, China, in 2022.

    His research interests include neural networks, computer vision, and anomaly detection.
\end{IEEEbiography}

\begin{IEEEbiography}[{\includegraphics[width=0.99in,height=1.25in,clip,trim=0cm 0.4cm 0cm 0.4cm]{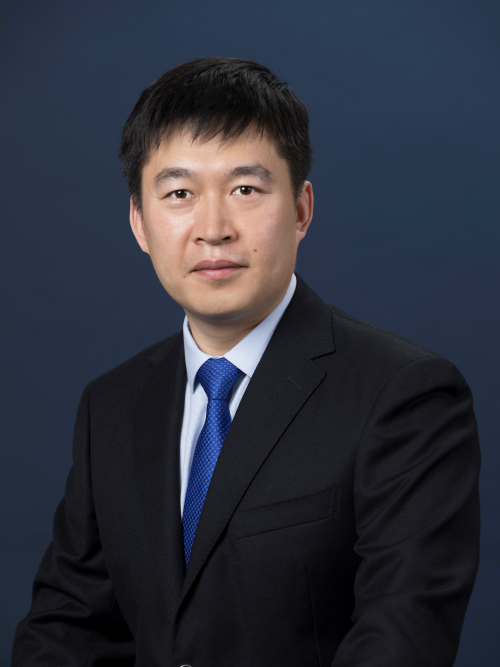}}]{Zhengtao Zhang}
    (Member, IEEE)
    received the B.S. degree from the China University of Petroleum, Dongying, China, in 2004, the M.S. degree
    from the Beijing Institute of Technology, Beijing, China, in 2007, and the Ph.D. degree
    in control science and engineering from the Institute of Automation, Chinese Academy of Sciences (IACAS), Beijing, in 2010.
    
    He is currently a Professor at IACAS and a Doctoral Supervisor at the School of Artificial Intelligence, University of Chinese Academy of Sciences.
    He also holds a part-time position as the Director at the Binzhou Institute of Technology, Binzhou, Shandong, China.
    His research interests include visual measurement, microassembly, and automation.
\end{IEEEbiography}

\vfill
\end{document}